\documentclass[10pt]{article} %

\usepackage[preprint]{rlj} %

\usepackage{amsmath,amsfonts,bm}

\def\eqref#1{equation~\ref{#1}}

\def\1{\bm{1}}

\DeclareMathAlphabet{\mathsfit}{\encodingdefault}{\sfdefault}{m}{sl}
\SetMathAlphabet{\mathsfit}{bold}{\encodingdefault}{\sfdefault}{bx}{n}

\DeclareMathOperator*{\argmax}{arg\,max}

 \usepackage{amssymb}            %
\usepackage{mathtools}          %
\usepackage{mathrsfs}           %
\usepackage{graphicx}           %
\usepackage{subcaption}         %
\usepackage[space]{grffile}     %
\usepackage{url}                %
\usepackage{lipsum}             %
\usepackage{booktabs}
\usepackage{placeins}
\usepackage[subtle]{savetrees}
\usepackage[ruled,vlined,linesnumbered]{algorithm2e}
\DontPrintSemicolon
\usepackage[capitalize,noabbrev]{cleveref}

\newif\ifdraft\drafttrue
\newif\ifaftersubmission\aftersubmissionfalse
\newif\iflater\latertrue %
\newif\ifjustbeforesubmission\justbeforesubmissionfalse  %
\newif\ifextended\extendedfalse

\definecolor{dkred}{rgb}{0.7,0,0}
\definecolor{dkpurple}{HTML}{4e02eb}
\definecolor{dkgreen}{HTML}{006329}
\definecolor{dkblue}{HTML}{2d5491}
\definecolor{dkorange}{HTML}{825a23}
\definecolor{ltgreen}{HTML}{3a9e67}
\definecolor{teal}{HTML}{007982}
\definecolor{fuchsia}{HTML}{8C368C}

\newcommand{\cwadd}[1]{#1}

\usepackage[addedmarkup=uline, defaultcolor=magenta, todonotes={textsize=scriptsize, textwidth=2.2cm}, authormarkuptext=name, commandnameprefix=always, xcolor]{changes} 

\definechangesauthor[name={JD}, color=orange]{jd}

\definechangesauthor[name={CW}, color=dkblue]{cw}

\def\taskreturn{cumulative task reward}

\def\waitreturn{cumulative waiting reward}

\def\lexq{LQ-Learning}
 \def\ldqn{LDQN}

\def\lextau{\sigma} %

\def\stS{\mathcal{S}}
\def\st{s}
\def\actS{\mathcal{A}}
\def\act{a}
\def\wait{\mathit{wait}}
\def\initDistr{\mu}
\def\dynamics{\mathcal{P}}

\def\basePolicy{\pi_{\mathit{base}}}
\def\actBase{\act_{\mathit{base}}}

\def\policy{\pi}

\newcommand{\distr}[1]{\Delta(#1)}

\def\barsp{\;|\;}

\title{Let it Cook:\\ Learning to Wait in Sequential Decision Making}

\setrunningtitle{Let it Cook: Learning to Wait in Sequential Decision Making}

\author{Christopher Watson, Arjun Krishna, Dinesh Jayaraman, Rajeev Alur}

\emails{\{ccwatson,arjk,dineshj,alur\}@seas.upenn.edu}

\affiliations{
\textbf{Department of Computer and Information Science, University of Pennsylvania, USA}
}

\contribution{
    We formalize learning to \textit{wait}---committing to take a distinguished \textit{wait} action that represents not actively pursuing task progress for a chosen duration while forgoing sensing---as a multi-objective optimization problem with lexicographically-ordered objectives in which the agent must wait as much as possible without sacrificing expected task performance.
    }
    {
    Prior work has studied the related problem of learning to commit to repeat arbitrary actions, but 
    \cwadd{use scalar objectives that implicitly define a tradeoff between task performance and action repetition.}
    }

\contribution{
    We propose to use a variant of Q-Learning with lexicographically-ordered objectives (\textit{\lexq{}}) to learn \textit{waiting policies} that sense and make a decision at as few timesteps as possible without sacrificing expected cumulative task reward. Additionally, this can be effectively used to train a \textit{wrapper waiting policy} that adapts a pre-trained policy to wait where appropriate.
    }
    {
    \cwadd{We extend an existing algorithm for lexicographic RL to handle extended-duration macro-actions.}
    }

\contribution{
    We show empirically that our approach can train ``waiting policies'' that (1) achieve task performance on par with that of non-waiting policies trained with vanilla RL and (2) perform extended-duration waiting where appropriate.
    }
    {
    Our experiments are performed in simulated environments.
    }


\summary{

In sequential decision making, an agent typically observes its environment and acts at every timestep.
However, such active participation may not always be necessary; tasks such as brewing coffee include periods that are served equally well by letting the environment evolve without constant monitoring and control.
During such periods, the agent could simply wait to conserve its resources, or redirect its attention to another task.
We capitalize on these opportunities by training 
a ``waiting policy'' that decides {\em where and how long to wait\/}. This involves forgoing sensing to commit to a \textit{wait} action, representing a deliberate pause for a set number of timesteps.
We formalize ``learning to wait'' as minimizing the frequency of sensing and decision making without sacrificing task performance (e.g., the total amount of time to complete a task).
To train a waiting policy, we propose an approach that employs reinforcement learning with lexicographically ordered objectives.
In experiments across $4$ discrete-state household tasks and $3$ continuous-state environments, we show that our approach successfully learns waiting behaviors, and can adapt pre-trained policies to wait where appropriate.
While different tasks permit different amounts of waiting without sacrificing task performance, our approach consistently finds solutions with significant waiting, sometimes waiting for over $50\%$ of the task duration.

 }

\begin{document}

\ifdefined\ARXIV
\else
\fi

\maketitle  %
\begin{abstract}

In sequential decision making, an agent typically observes its environment and acts at every timestep.
However, such active participation may not always be necessary; tasks such as brewing coffee include periods that are served equally well by letting the environment evolve without constant monitoring and control.
During such periods, the agent could simply wait to conserve its resources, or redirect its attention to another task.
We capitalize on these opportunities by training 
a ``waiting policy'' that decides {\em where and how long to wait\/}. This involves forgoing sensing to commit to a \textit{wait} action, representing a deliberate pause for a set number of timesteps.
We formalize ``learning to wait'' as minimizing the frequency of sensing and decision making without sacrificing task performance (e.g., the total amount of time to complete a task).
To train a waiting policy, we propose an approach that employs reinforcement learning with lexicographically ordered objectives.
In experiments across $4$ discrete-state household tasks and $3$ continuous-state environments, we show that our approach successfully learns waiting behaviors, and can adapt pre-trained policies to wait where appropriate.
While different tasks permit different amounts of waiting without sacrificing task performance, our approach consistently finds solutions with significant waiting, sometimes waiting for over $50\%$ of the task duration.

 \end{abstract}

\section{Introduction}\label{sec:introduction}

The prevailing paradigm in sequential decision making treats agents as active, closed-loop systems that continually monitor and react to their environment. This approach assumes that making meaningful progress towards a goal requires constant engagement from the agent---resulting in active utilization of its sensory, computational, and motor resources. However, many real-world tasks inherently possess passive dynamics that can be exploited to achieve goals more efficiently~\citep{todorov2009efficient}. Consider the mundane but complex choreography of a busy %
kitchen: 
a kettle takes time to boil, a soup takes time to cook, and a coffee machine operates independently once set to brew.
In each of these scenarios, the most effective action for a human, or a robot, is to simply wait.
An agent that identifies where and how long to wait can switch to a timed ``standby mode'' to conserve cognitive and motor resources, or alternatively redirect these resources to an auxiliary task.
Given the benefits of strategically waiting, the key questions that arise are: 
How would one 
(1) formalize the objective of waiting? and (2) develop learning approaches that train an agent to optimize this waiting objective?

For each environment, we define a distinguished \textit{wait} action, e.g., standing still while coffee brews, or holding the paddle still in the video game of \textit{Pong}. Then "waiting" is simply the act of committing to perform this \textit{wait} action for multiple consecutive timesteps.
We aim to discover and exploit opportunities where the agent can wait for an extended period.
While our approach is technically agnostic to the effect the \textit{wait} action has on the environment, we limit our experiments to settings where ``waiting'' corresponds to letting the environment evolve according to an intuitive notion of passive dynamics, as in the aforementioned examples.

Learning to wait balances two potentially competing %
\cwadd{objectives}: maximizing task performance and minimizing
the number of times the agent has to sense and respond to the environment 
(which corresponds to maximizing the amount of time the agent spends waiting).
We consider agents that strictly prioritize task performance 
(measured by cumulative reward from a Markovian environment)
and only seek to maximize waiting when doing so does not degrade task performance.
Formally, this preference transforms the problem into a multi-objective optimization problem with a lexicographic ordering: the agent seeks to find $\max\:\: [J^0, J^1]$, where the \textit{\taskreturn{}} $J^0$ is represents expected task performance and \textit{\waitreturn{}} $J^1=-D$ decreases with the number of times $D$ the agent is called upon to sense and make a decision (to either wait or act).
 For us, waiting periods involve no sensing, no computation, and no task-oriented behavior. Minimizing $D$ thus combines two desiderata: (1) reducing sensing and decision computation, and (2) reducing the need for the agent's active interventions.
 In a multi-tasking setting, waiting may correspond to periods during which the agent could redirect its resources towards another task.
 More generally, waiting periods permit switching to a timed ``standby mode'' that conserves sensory, computational, and motor resources.

To exploit our objective's structure, we apply a \textit{lexicographic} Multi-Objective Reinforcement Learning~(MORL) algorithm across several discrete- and continuous-state environments. 
The amount of waiting possible depends on the environment; in some environments, our method yields policies that wait (in some environments, for over 50\% of the task duration) without sacrificing task performance \cwadd{compared to a vanilla RL, non-waiting baseline.}
We further compare our method against a standard MORL technique that uses scalarized rewards to explore the Pareto frontier between task performance and increased waiting, mediated by a reward-weighting scalarization coefficent.
Under a strict preference for task performance, our lexicographic MORL approach achieves 
\cwadd{near-identical task and waiting performance to the scalarized approach, without needing a hyperparameter sweep to discover the best reward-weighting coefficient for each environment.}
\cwadd{We also show that our approach can be used to \textit{wrap} a pretrained policy to add waiting behavior without sacrificing task performance.}
Finally, we demonstrate how learned waiting behavior may enable policy interleaving in a multi-task scenario.

\section{Waiting in Sequential Decision Making}\label{sec:setting}

\begin{figure*}[th]
  \centering
\includegraphics[width=\textwidth]{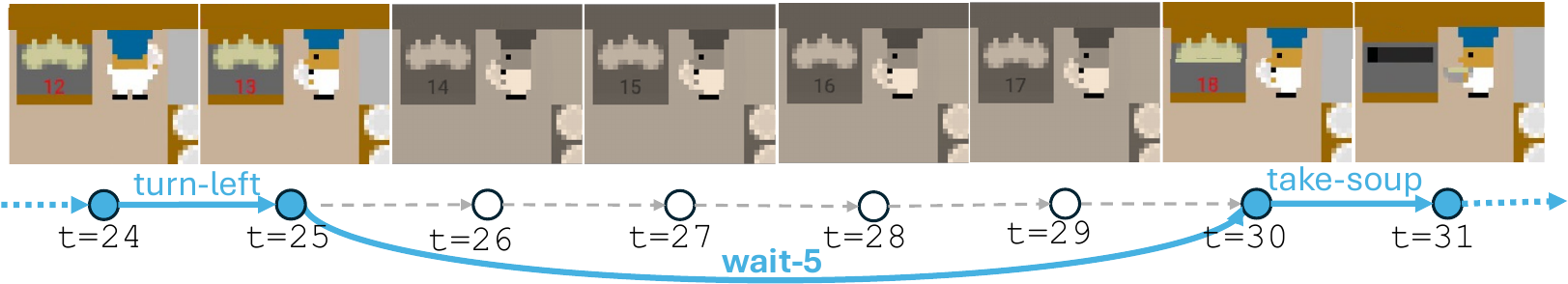}
\caption{
In our \texttt{Cook} environment, the agent commits to a duration-5 waiting macro-action near the end of the soup cooking process. No sensing, computation, or active motion occurs while waiting.}
  \label{fig:concept}
\end{figure*}

We define a \textit{waiting Markov Decision Process (WMDP)} to be a tuple $\mathcal{M} = (\stS, \actS, \wait, \mathcal{W}, \dynamics, R, \initDistr)$
where $\stS$ is the (discrete or continuous) set of states, 
$\actS$ is the (discrete) set of primitive actions, 
$\wait \in \actS$ is the distinguished \textit{wait} action,
$\mathcal{W} \subseteq \mathbb{N}^+$ is the (finite) set of waiting durations,
\cwadd{$\dynamics(\st_{t+1} \barsp \st_t, \act_t)$ is the probability of transitioning from $\st_t$ to $\st_{t+1}$ upon taking primitive action $\act_t$,}
$R: \stS \times \actS \times \stS \rightarrow \mathbb{R}$ is the reward function, 
and $\initDistr$ is the initial state distribution.
What differentiates a WMDP from a standard Markov decision process is the distinguished $\wait$ action and the set $\mathcal{W}$ of waiting durations.
Each natural number $w \in \mathcal{W}$ characterizes a \textit{waiting macro-action} that corresponds to waiting for $w$ consecutive timesteps---that is, forgoing sensing and executing the primitive $\wait$ action for $w$ consecutive timesteps.\footnote{A \textit{waiting} macro-action is a special kind of \textit{semi-Markov option} in the terminology of~\citet{sutton1999between}.}
We will denote the duration-$w$ waiting macro-action by its natural number $w$.
We do not make any special requirements of the dynamics $\dynamics$ under the \textit{wait} action, however in applications we will choose the \textit{wait} action in a way that corresponds to a natural notion of non-intervention (e.g. standing idly while soup boils on the stove, or not moving the paddle in a game of \textit{Pong}).
We aim to learn policies that can commit to ``wait'' for an extended period, as illustrated in~\cref{fig:concept}.

The $\wait$ action and set of waiting durations $\mathcal{W}$ let us define the notion of a \textit{waiting policy}
${\policy : \stS \rightarrow \distr{\actS \cup \mathcal{W}}}$
that interacts with the WMDP to receive both \textit{task rewards} and \textit{waiting rewards}.\footnote{For a measurable space $\mathcal{X}$, we write $\distr{\mathcal{X}}$ to denote the set of probability measures over $\mathcal{X}$.}
When queried at a state $\st$, the waiting policy yields either a primitive action $\act \in \actS$, which has its usual effect according to the dynamics $\dynamics$, or a waiting macro-action $w \in \mathcal{W}$. %
A WMDP $\mathcal{M} = (\stS, \actS, \wait, \mathcal{W}, \mathcal{P}, R, \mu)$, a waiting policy $\policy : \stS \rightarrow \distr{\actS  \cup \mathcal{W}}$, and a finite time horizon $H \in \mathbb{N}$ induce a distribution over length-$H$ trajectories 
$\st_0 \act_0 r^0_1r^1_1\st_1\ldots \act_{H-1}r^0_Hr^1_H\st_H$ generated by drawing an initial state $\st_0 \sim \mu$ and querying the policy $\pi$ at timestep $0$ to obtain either a waiting macro-action or a primitive action.
At each timestep $t$, if the policy is queried and yields a waiting macro-action $w$ then the next $w$ actions $\act_t, \act_{t+1}, \ldots, \act_{t+w-1}$ are each the waiting action $\wait$ and the policy is not queried again until timestep $t+w$. 
If instead the the queried policy yields a primitive action $\act \in \actS$ then $\act_t=\act$ and the policy is queried at the next timestep $t+1$.
For each timestep $1 \le t \le H$, the
\cwadd{state $\st_t$ is sampled as  $\st_t {\sim} \dynamics(\cdot \barsp \st_{t-1}, \act_{t-1})$} and the \textit{task reward} $r^0_t$ is defined as $r^0_t = R(\st_{t-1}, \act_{t-1}, \st_t)$, which matches the usual notion of reward in an MDP. 
The \textit{waiting reward} $r^1_t$ is $-1$ if the policy is queried at timestep $t$ and $0$ otherwise.
When the maximum episode horizon $H$ is reached, the episode is abruptly terminated, regardless of whether the agent is in the middle of a waiting macro-action.

Given a trajectory 
$\tau = \st_0 \act_0 r^0_1r^1_1\st_1\ldots \act_{H-1}r^0_Hr^1_H\st_H$
we define the (undiscounted) \textit{\taskreturn{}} to be $J^0(\tau)= \sum_{1 \le i \le H}r^0_i$.
We define the (undiscounted) \textit{\waitreturn{}} to be $J^1(\tau) = \sum_{1 \le i \le H}r^1_i$, i.e., $-1$ times the number of times the policy $\pi$ is queried during the trajectory.

\paragraph{Lexicographic waiting objective.}
Intuitively, a good waiting policy is one that obtains good task performance (as measured by \taskreturn{}) while also waiting as much as possible.
Since we consider trajectories with an \textit{a priori} fixed horizon $H$ in which the only extended-duration actions are the \textit{waiting macro-actions}, querying the waiting policy fewer times corresponds to more waiting.
Given a WMDP $\mathcal{M}$ and a finite time horizon $H$, we define the vector-valued objective of a waiting policy $\pi$ as $J(\pi) \in \mathbb{R}^2$ as:
\begin{equation}\label{eqn:objective}
    J(\pi) = \mathbb{E}_{\tau \sim \mathcal{M}, \pi}[J^0(\tau),J^1(\tau)]
\end{equation}
We adopt the standard lexicographic ordering $\le$ over objective values defined such that $[{J^0}, {J^{1}}] \le [{J^{0}}', {J^{1}}'] \Leftrightarrow {J^{0}} < {J^{0}}' \,\text{OR} \,({J^{0}} = {J^{0}}'\, \text{AND} \,{J^{1}} \leq {J^{1}}')$. In other words, a policy is ``better'' than another policy if and only if it either (1) achieves higher \taskreturn{} $J^0$, or (2) it achieves equal $J^0$ and higher \waitreturn{} $J^1$.
In the following section we will describe our RL-based approach that seeks to learn a policy $\pi^*$ that achieves maximal $J(\pi^*)$.

Our choice of waiting objective---to minimize expected number of times the waiting policy is queried---is a natural fit for settings where active decision making incurs a computational or sensory cost.
\cwadd{This idea was explored by~\citet{zhou2024timing}, which uses scalar rewards that include an \textit{interaction cost} to encourage action repetition, and~\citet{harb2018when}, which uses the analogous \textit{deliberation cost} to reduce the number of times a high-level policy is queried in hierarchical reinforcement learning.}
In the aforementioned works, the magnitude of the (interaction or deliberation) cost implicitly defines a tradeoff between task performance and policy query frequency.
In contrast, our lexicographic objective~(\ref{eqn:objective}) renders the relative magnitudes of the task rewards and the waiting rewards inconsequential.
In other words, if we were to define WMDP trajectories such that each $r^1_t$ were scaled by an arbitrary positive factor, the objective in~(\ref{eqn:objective}) would induce the same ordering over optimal policies and our \lexq{} learning algorithm (\cref{alg:lexq}, which we introduce in~\cref{sec:lexicographic}) would behave similarly.
Practically, this alleviates the need to perform a hyperparameter sweep to find an appropriate weighting of task reward vs.~waiting reward.

Another important distinguishing characteristic of our formalization is that
the only way to avoid querying the policy is by selecting the temporally extended waiting macro-actions. Thus, our objective naturally encourages the agent to pick those actions. Further, since ``wait'' actions in our settings cede control to the environment's passive dynamics, more waiting translates to conserving  not only sensing and computation, but also motor resources. 
Penalizing all decisions, including the choice to commit to waiting, encourages the agent to take \textit{longer duration} waiting actions when possible. 
Preferring long contiguous durations of waiting can enable downstream multitasking applications where a long waiting period provides more time for the agent to ``fill in'' the waiting period by executing an auxiliary policy to make progress towards another task;
we explore a simple instantiation of this idea in the context of the \texttt{Coffee} task in~\Cref{sec:experiments}.

\section{Learning to Wait}\label{sec:approach}

\subsection{Lexicographic Q-Learning}\label{sec:lexicographic}

Our goal, to maximize the vector-valued objective~(\ref{eqn:objective}), lends itself naturally to a \textit{lexicographic} MORL approach~\citep{gabor1998multi,skalse2022lexicographic} that seeks a policy that maximizes a vector-valued reward signal.
We adapt the lexicographic version of Q-Learning described by~\citet{skalse2022lexicographic} to handle waiting macro-actions. %
Our resulting \lexq{} algorithm, which we detail in \cref{alg:lexq}, maintains two Q-estimates: the task Q-estimate $Q^0$ and waiting Q-estimate $Q^1$.
Each Q-estimate considers the ``full action set'' $\actS \cup \mathcal{W}$, which contains both primitive actions and waiting macro-actions.
Greedy policy inference 
selects an action that is $\lextau$-close to optimal with respect to $Q^0$'s future cumulative task reward estimate, breaking ties according to $Q^1$'s future cumulative waiting reward estimate, where $\lextau \in \mathbb{R}^+$ is a small tolerance parameter that softens the lexicographic comparison.
The tolerance parameter prevents a slightly inaccurate empirical $Q^0$ estimate from erroneously ruling out the lexicographically optimal action during the $Q^1$ update in~\Cref{line:q1-update} of~\cref{alg:lexq}.
We only update the Q estimates in response to the output (an element of $\actS \cup \mathcal{W}$) made by the policy when the policy is queried. We do not perform updates based on the primitive $\wait$ actions applied in the middle of a waiting macro-action, because these do not directly correspond to decisions made by the policy. This is analogous to how~\citet{sutton1999between} defines \textit{SMDP Q-Learning}.

\begin{algorithm}[th]
\caption{\lexq{}}\label{alg:lexq}
\DontPrintSemicolon
\SetKwInput{KwIn}{Input}
\SetKwProg{Fn}{Function}{:}{}
\SetKwFor{ForEach}{for each}{do}{end}
\SetKw{KwRet}{return}

\KwIn{WMDP $\mathcal{M} = (\stS,\actS,\wait,\mathcal{W},\dynamics, R, \initDistr)$, episode horizon $H$, learning rate $\alpha$, lexicographic tolerance $\lextau$, initial Q tables $Q^0, Q^1 \in \mathbb{R}^{|\stS|\times|\actS\cup\mathcal{W}|}$.}

\ForEach{episode}{
  $t \gets 0, \; \st \gets \mathcal{M}.\texttt{reset()}$\; 

  \While{$t < H$}{
    $u \gets 
\begin{cases}
\texttt{Lex-Argmax}(\st, Q^0, Q^1, \lextau) & \text{w.p. } 1-\epsilon \\
\mathrm{Uniform}(\actS \cup \mathcal{W}) & \text{w.p. } \epsilon
\end{cases}$\tcp*[f]{epsilon-greedy exploration}\;

    \uIf(\tcp*[f]{Waiting macro-action}){$u \in \mathcal{W}$}{
      $G^0 \gets 0$\tcp*[f]{Task reward during macro-action}\;
      \For(\tcp*[f]{Take u primitive wait steps}){$u$ times}{
        $\st', r^0 \gets \mathcal{M}.\texttt{step}(\wait)$\;
        $G^0 \gets G^0 + r^0$\;
      }
    }\Else(\tcp*[f]{Primitive action}){

      $\st', r^0 \gets \mathcal{M}.\texttt{step}(u)$\;
      $G^0 \gets r^0$\;
    }

    $G^1 \gets -1$\tcp*[f]{Waiting reward: -1 per policy query}\;

    \tcp{Update Q estimates}
    $Q^0[\st,u] \gets (1-\alpha)\,Q^0[\st,u] + \alpha\Bigl(G^0 + \max_{v\in\actS\cup\mathcal{W}} Q^0[\st',v]\Bigr)$\;

    $Q^1[\st,u] \gets (1-\alpha)\,Q^1[\st,u] + \alpha\Bigl(G^1 + Q^1[\st',\texttt{Lex-Argmax}(\st', Q^0, Q^1, \lextau)]\Bigr)$\;\label{line:q1-update}
    $\st \gets \st', \; t \gets t + 1 $\;
  }
}
\vspace{0.25em}
\Fn{\texttt{Lex-Argmax}$(\st, Q^0, Q^1, \lextau)$}{
  $U \gets \Bigl\{u \in \actS \cup \mathcal{W}\ \Big|\ Q^0[\st,u] \ge \max_{u'\in\actS\cup\mathcal{W}} Q^0[\st,u'] - \lextau \Bigr\}$\;
  \KwRet $\argmax_{u\in U} Q^1[\st,u]$\;
}

\end{algorithm}

\subsection{Learning A Waiting Wrapper For a Pretrained Policy}\label{sec:adapting}
Thus far, we have focused on training a policy \textit{from scratch}.
In some settings, there may already exist a policy $\basePolicy : \stS \rightarrow \distr{\actS}$ that achieves satisfactory cumulative (task) reward in a non-waiting MDP environment. 
Such a policy may be handwritten, or trained via e.g. imitation \cwadd{learning} or RL.
Direct deployment of $\basePolicy$ in a WMDP\footnote{Deploying $\basePolicy: \stS {\rightarrow} \distr{\actS}$ in a WMDP 
with actions $\actS$ and wait durations $\mathcal{W}$
assumes the tacit lift to  $\basePolicy : \stS \rightarrow \distr{\actS \cup \mathcal{W}}$.}
would garner expected \taskreturn{} $J^0(\basePolicy)$ equal to expected cumulative reward in a non-waiting MDP, but extremely low \waitreturn{}.
We will now describe how to train a waiting policy that serves as a wrapper around $\basePolicy$.
This approach enjoys efficient exploration because the wrapper policy has a small set of possible actions: either execute $\basePolicy$ to evolve the system under its controlled dynamics, or wait and let it evolve passively.

Given a WMDP $\mathcal{M} = (\stS, \actS, \wait, \dynamics, R, \initDistr)$ and a base policy $\basePolicy : \stS \rightarrow \distr{\actS}$ we define the induced two-action WMDP $\mathcal{M}' = (\stS, \{\actBase, \wait\}, \dynamics', R, \initDistr)$ where $\actBase \notin \actS$ is a new primitive action that ``simulates'' $\basePolicy$.
The dynamics are defined such that $\forall \st, \st' \in \stS$\begin{equation*}
    \dynamics'(\st' \barsp \st, \wait) = \dynamics(\st' \barsp \st, \wait) \qquad \qquad
    \dynamics'(\st' \barsp \st, \actBase) =\sum_{a \in \actS}\left( \dynamics(\st' \barsp \st, \act)\basePolicy(\act \barsp \st) \right)
\end{equation*}
A waiting policy $\pi : \stS \rightarrow \distr{\{\actBase, \wait \} \cup \mathcal{W}}$ can choose to simulate $\basePolicy$ by taking action $\actBase$ or wait.
Learning a waiting wrapper around a pre-trained policy $\basePolicy$ may be useful in several settings. 
When training waiting policies from scratch requires extensive environmental interaction then taking advantage of an existing base policy $\basePolicy$ (trained without waiting considerations) may significantly reduce the amount of environmental interaction needed to obtain a policy that achieves satisfactory value of~\cref{eqn:objective}.
For example, in domains where large pre-trained policies exist, such as in robotic manipulation,
(e.g. \citet{octo2023octo,brohan2023rt2,black2026pi0,gr00tn1_2025})
it may be unnecessary to train a policy from scratch.
Using the method described in this section, we may
exploit the pre-trained base policy's task performance while also exploring its waiting affordances for improved sensory, computational, and motor resource efficiency during task execution.
Of course, it is possible that there exists no wrapper waiting policy $\pi$ with expected objective value that exceeds 
$J(\st \mapsto \delta_{\actBase})$.
In our experimental results, we observed non-trivial waiting behaviors even when the base policy $\basePolicy$ was trained through standard RL (with no additional care to  make it robust in states that may be reached via extended-duration waiting).

\section{Experiments}\label{sec:experiments}

We compare the effectiveness 
of \cwadd{our \lexq{} approach}
(\cref{sec:lexicographic}) and 
\cwadd{a scalarized reward baseline}
 (which we will describe in \cref{sec:scalar}), both in the context of learning a waiting policy from scratch and wrapping a pre-existing policy to wait as much as possible without sacrificing expected \taskreturn{}.
Finally, we show a case study of how learning to wait can be used as a first step toward interleaving policies to complete multiple tasks efficiently. 

\subsection{Environments}

Each of our environments (with the exception of \texttt{CartPole}) is a goal-reaching environment: we consider the task reward to be $-1$ per timestep, which encourages the agent to reach the goal as fast as possible. 
We also include \textit{early episode termination} if the agent reaches the goal
before the \textit{a priori} episode horizon is reached.
While our choice of task reward allows us to visualize \taskreturn{} and the number of timesteps in which the policy is queried (i.e., zero minus the \waitreturn{}) on a single set of axes, we note that our approach also applies to settings where \taskreturn{} does not enjoy any special correlation to a number of elapsed timesteps.
Our environments are 
described below, with additional details in~\cref{app:env-details}.
\cwadd{In all of our environments the agent observes the complete state of the environment; in the cooking and coffee preparation tasks this includes the amount of timesteps remaining for a soup to finish cooking, the coffee machine to finish heating, or the coffee to finish brewing.}

\paragraph{Cook.} We study three variations of a cooking task in the \texttt{overcooked-ai}~\citep{caroll2019utility}, which simulates the popular video game \textit{Overcooked}~\citep{overcooked} in which the agent must cook and serve dishes.
In our \texttt{Cook} environment the agent must gather and place 3 onions in the pot, which starts an 18 timestep countdown until the soup is cooked and ready. The agent must also pick up a dish, use the dish to pick up the soup, and then deliver the soup to the goal location.
In \texttt{CookLonger} the soup takes 36 timesteps to cook, and in \texttt{CookTwice}, the agent must cook and deliver two soups, each of which takes 18 timesteps to cook. 
The $\wait$ action makes the agent stand still and we consider $\mathcal{W} = \{5, 15\}$. 

\paragraph{Coffee.} We implement a \texttt{Coffee} brewing task in a deterministic 2D MiniGrid gridworld~\citep{chevalier2023minigrid}. The agent must go to the coffee machine and \textit{toggle} it to start it heating. After an 8 timestep countdown the agent must toggle the machine again to start coffee brewing, which triggers an 18 timestep countdown, after which the agent must toggle the coffee machine again to collect the coffee.
In addition to the multi-phase interaction with the coffee machine, the agent must go to and \textit{toggle} the cream and sugar to pick them up.
The episode terminates (success) when the coffee, cream, and sugar are collected (in any order).
The $\wait$ action makes the agent stand still and we consider $\mathcal{W} = \{5, 15\}$.

\paragraph{Pong.} Our \texttt{Pong} environment is a modified version of the Atari game Pong, rewritten with inspiration from the Gymnax~\citep{gymnax2022github} implementation. The agent plays against a scripted opponent, and its objective is to win one point as quickly as possible.
If the ego agent loses a point, the episode does not terminate but the ball restarts in the middle of the arena.
The agent observes a continuous-valued low-dimensional vector observation comprising each paddle's position, the ball's current and last timestep positions, and the ball's current velocity.
The $\wait$ action does not move the paddle and we consider $\mathcal{W} = \{2, 4, 8, 16, 32, 64, 128\}$.

\paragraph{MountainCar.} We adopt Gymnasium's~\citep{towers2024gymnasium}  \texttt{MountainCar}, in which 
a car must climb up from a valley to reach the top of a mountain as fast as possible. This is only possible by gradually building up momentum over multiple back-and-forth passes. Unlike our other environments, even small amounts of waiting hinder task performance.
The agent observes the car's position and velocity.
The $\wait$ action applies zero force to the car (which may still roll freely) and we consider $\mathcal{W} = \{2, 4, 6, 8, 10, 12, 14, 16, 18, 20\}$.

\paragraph{Cartpole.} We adapt Gymnasium's~\citep{towers2024gymnasium} \texttt{Cartpole} environment.
We add an additional primitive action $\wait$ to standard CartPole's 2-action set (``left'' and ``right''). Unlike our other environments, the goal is to balance the pole as long as possible, to the maximum horizon $H=200$; we use a task reward of $+1$ per timestep.
The agent observes the cart's position and velocity, and the pole's angle and angular velocity.
We consider $\mathcal{W} = \{2, 4, 6, 8, 10, 12, 14, 16, 18, 20\}$. 

\subsection{Scalar reward baseline}\label{sec:scalar}
\cwadd{A simpler alternative to \lexq{} uses a coefficient $\lambda$ to scalarize each task reward $r^0$ and waiting reward $r^1$ into a combined reward $r^0 + \lambda r^1$.
To train a policy that optimizes~\cref{eqn:objective}, one must perform a hyperparameter sweep over values of $\lambda$ in each new environment.
Our \lexq{} approach does not require such hyperparameter tuning, however lexicographic MORL approaches (such as \lexq{}) require specific policy architectures and algorithms (e.g. those presented in~\citet{skalse2022lexicographic}) that are less mature than scalar-reward RL approaches.}
For a scalarization coefficient $\lambda \in \mathbb{R}^{\ge 0}$ we define the $\lambda$-scalarized expected cumulative reward of a policy (for a fixed waiting MDP and horizon) to be
    $J_{\lambda}(\pi) = J^0(\pi) + \lambda J^1(\pi)$
In order to train a policy that optimizes $J_\lambda$ we apply the non-lexicographic version of the SMDP Q-Learning algorithm~\citep{sutton1999between} defined similarly to our \lexq{} (\cref{sec:lexicographic}) except there is only one Q-estimate and the (scalarized) reward at timestep $t$ is $r_t = r^0_t + \lambda r^1_t$.
When ${\lambda=0}$  we recover the original task reward's objective that does not incentivize waiting.
When $\lambda$ is extremely large, there will always be an optimal policy that waits as much as possible, regardless of the task rewards.%

\subsection{Implementation details}
We apply \lexq{} and Q-Learning with scalar rewards to the discrete \texttt{Cook}, \texttt{CookTwice}, \texttt{CookLonger} and \texttt{Coffee} tasks. 
We apply their deep RL analogs \ldqn{} (details in~\cref{app:ldqn}) and DQN~\citep{mnih2015human} to the continuous-state environments. 
We adapt the Q-Learning and DQN implementations found in SKRL~\citep{serrano2023skrl} to \lexq{} and \ldqn{} following~\citet{skalse2022lexicographic}.
Our DQN-based experiments use Q-networks that are MLPs with ReLU activations and 2 hidden 256 neurons per layer. Further hyperparameter details can be found in~\cref{app:training}.
For the scalar reward baseline, we sweep over the $\lambda$ values $0,0.01, 0.1, 0.5, 1, 5, 10, 15$.
We report the result of the best $\lambda$ value with respect to~\Cref{eqn:objective}, additional results are found in~\cref{app:extended-results}.

\subsection{Results}

\begin{figure}[!h]

    \centering
        \includegraphics[width=\columnwidth]{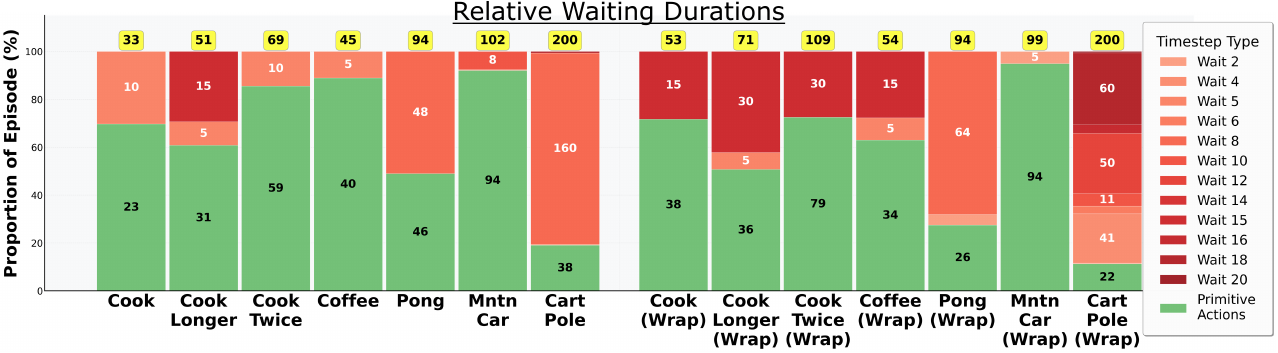}
    \caption{
    Empirical performance averaged over 1000 trajectories from the best trained policy of 10 random training seeds using \lexq{} or \ldqn{}. The bar segments show the proportion of timesteps spent waiting vs. actively engaging with the environment. Bar heights are normalized; the numbers within each segment denote the absolute number of timesteps and the number at the top denotes absolute episode length.
    }
    \label{fig:scratch-and-adapt}
\end{figure}

\begin{figure}[!h]
    \centering

        \includegraphics[width=\columnwidth]{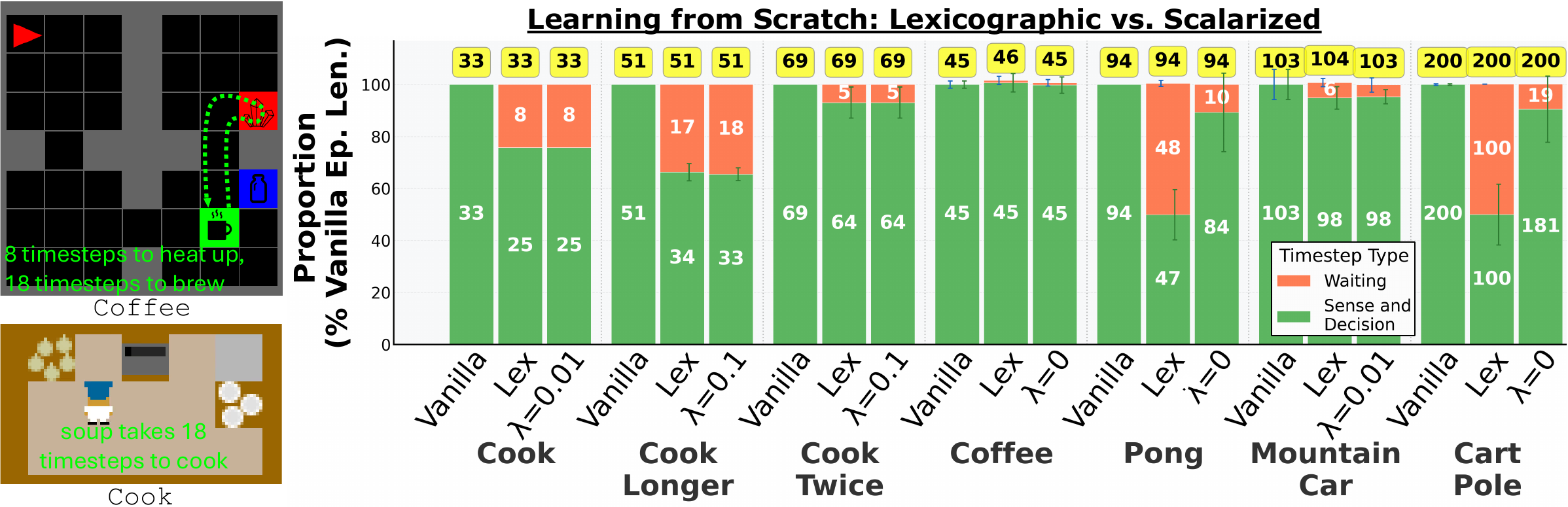}
    \caption{
    (\textbf{Left}) Environment visualizations. 
    In our \texttt{Coffee} interleaving case study, the duration-15 wait taken while coffee brews permits us to interleave a ``fetch sugar'' policy that follows the dashed green path. 
    (\textbf{Right}) Empirical mean episode length (height of bar) and  number of decisions made (height of green bar segment) of our approach (``Lex'' denotes \lexq{} for discrete environments, \ldqn{} for others) and the best scalarized approach from our sweep over $\lambda$ values. Episode length is negative cumulative task reward for most environments; for \texttt{CartPole} episode length is cumulative task reward. We also include results from vanilla RL (Q-Learning for discrete environments, DQN for others) that is trained on task reward without the option to wait. Within each environment, heights are normalized with respect to the vanilla RL baseline. Labels denote absolute numbers of timesteps. Error bars represents $\pm$ 1 standard deviation, computed over 10 random initial policy training seeds.
    }
    \label{fig:coffee}
    \label{fig:cluster}
\end{figure}

\paragraph{To what extent can an agent wait?}
We first study the extent to which an agent can wait without sacrificing task performance across our suite of tasks. 
We apply our \lexq{}/\ldqn{} approach (\Cref{sec:lexicographic}). %
Our comparison in~\Cref{fig:cluster} shows that 
\cwadd{\lexq{} and \ldqn{} consistently yield} policies that achieve high \textit{cumulative task reward} (close to that obtained by vanilla RL that maximizes \taskreturn{} as its sole objective without the option to wait), so a quick summary of the waiting affordances of the task exploited by our approach is possible through~\Cref{fig:scratch-and-adapt}, which plots the proportion of episode timesteps spent in different duration waiting macro-action by our \lexq{} and \ldqn{} policies. 

Intuitively, \texttt{Cook} and \texttt{Coffee} (visualized in \cref{fig:coffee}) would present opportunities for the agent to wait (e.g. during the time that soup takes to cook or coffee takes to brew) without sacrificing task performance (here, time to goal).
Indeed, in \texttt{Cook} the learned policy chooses the duration $5$ wait action twice during the 33 timesteps taken to fetch ingredients, cook, and serve the soup.
At first it seems counterintuitive that the agent does not take the duration $15$ waiting action given that the soup takes $18$ timesteps to cook, however our \lexq{}-learned policy is in fact optimal with respect to \cref{eqn:objective}.
Close inspection of the learned policy's behavior reveals that our \lexq{} approach discovers a policy that fetches the dish during the time the soup is cooking: this permits optimal \taskreturn{}, and an optimal value of~\Cref{eqn:objective}, yet precludes the opportunity to wait for 15 uninterrupted timesteps.\footnote{Any policy that performs a duration-15 waiting macro action would take at least 37 timesteps to complete \texttt{Cook}.}
Similarly, in \texttt{Coffee}, \lexq{} learns a clever policy that achieves a higher value of~\cref{eqn:objective} than possible by naively waiting: the learned policy fetches sugar as the coffee brews.

\cwadd{Training a waiting wrapper policy around a pre-existing policy (\Cref{sec:adapting}) can expose different opportunities to wait without sacrificing task reward than when learning a waiting policy from scratch.}
\cwadd{We see this when we train a waiting wrapper policy to wrap a (suboptimal) handwritten policy in our tabular environments:}
Our naive handwritten \texttt{Cook} base policy stands idly by as the soup cooks; our wrapper waiting policy  trained via \lexq{} successfully learns to take the duration-15 wait during this idle period.
Similarly, our wrapper waiting policy for \texttt{Coffee} performs a duration-5 and a duration-15 wait as the machine heats and brews, respectively.
In our continuous-state environments, we wrap a policy trained using vanilla DQN.
Surprisingly, even these natural RL-trained afford the ability to exercise waiting behaviors. For example, our learned waiting wrapper policies for \texttt{Pong} and \texttt{Cartpole} preserve the DQN-trained base policy's perfect cumulative task reward while spending well over half of the timesteps committed to waiting. This suggests that learning to wait, conveniently applied as a wrapper \textit{over} pre-trained policies, could query the base policy more selectively without reducing task performance.%

\paragraph{\lexq{} is a good way to learn waiting (compared to the scalar reward baseline).}

We now investigate the extent to which our \lexq{}/\ldqn{} approach accurately finds a policy that toes the fine line between increasing waiting and sacrificing task performance. 
\Cref{fig:cluster} visualizes episode duration (which directly reflects \taskreturn{}) and number of decisions made (zero minus \waitreturn{}) for three approaches for each task: (1.) Vanilla RL that maximizes task performance with no access to waiting options, (2.) Our \lexq{}/\ldqn{} approach, 
and (3.) The scalarized objective approach with the best (as determined by \Cref{eqn:objective}) weighting coefficient found in our sweep over $\lambda$ values.
A policy that maximizes~\Cref{eqn:objective} should yield episode length approximately equal to that achieved by Vanilla RL, and should wait as much or more than the scalarized approach.
Our \lexq{} and \ldqn{} policies in \Cref{fig:cluster} indeed approximate this expected behavior.

\begin{figure}[h]
    \centering
    \includegraphics[width=\columnwidth]{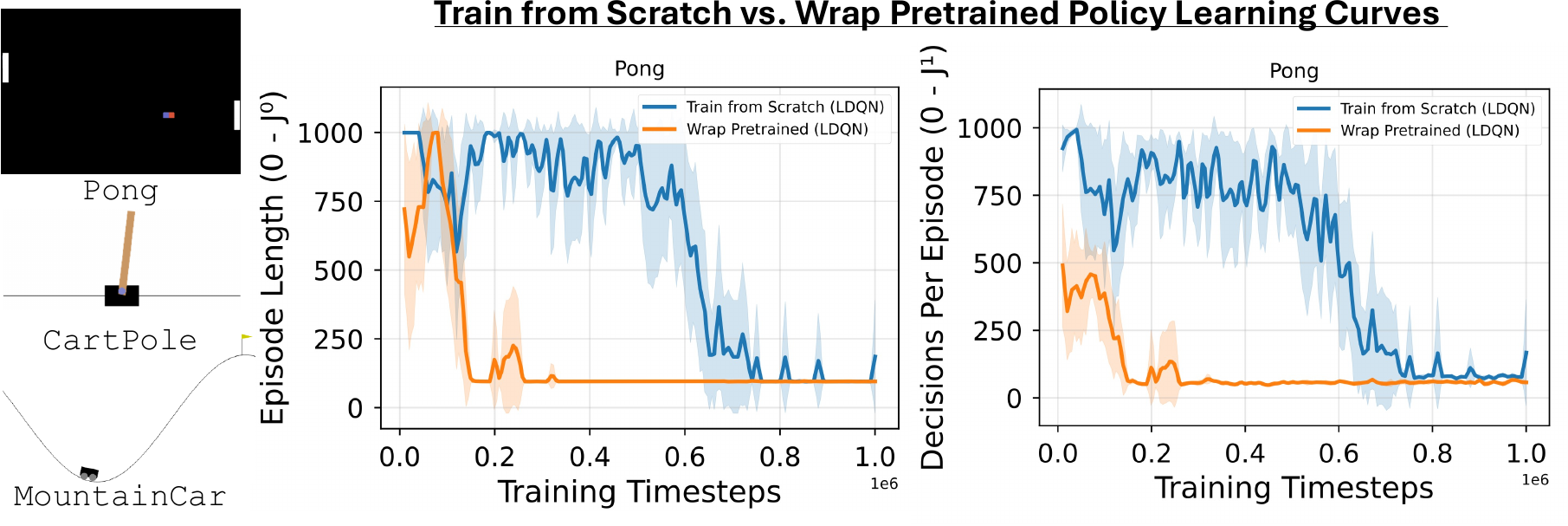}

    \caption{%
    \textbf{(Left)} Environment visualizations. \textbf{(Right)} Learning curve comparison for our \ldqn{} applied to train a waiting policy for \texttt{Pong} from scratch vs. as a wrapper around a frozen pretrained policy. Each curve is the mean of 10 random policy training seeds, shaded regions denote 95\% confidence interval.
    }
    \label{fig:pong-curves}
\end{figure}

\paragraph{Learning a waiting policy wrapper is sample efficient.}
We find it often takes fewer training steps to saturate the performance of a wrapper waiting policy than a waiting policy trained from scratch.
For example, in \texttt{Pong} training a waiting wrapper policy via \ldqn{} converges after ${\sim}250\text{k}$ training environment steps, while training a waiting policy from scratch with \ldqn{} typically takes more than $600\text{k}$ steps (\Cref{fig:pong-curves}).
This gain in sample efficiency may be due to (1) fewer actions being exposed to the wrapper policy and (2) a stark dichotomy between states from which the base policy can perform well vs. poorly, which means the waiting wrapper consistently receives low (relative to not waiting) future cumulative task reward when it waits inappropriately.
Such a gain in sample efficiency need not exist in general, however learning curves in~\Cref{app:curves} show efficiency gains in each environment except \texttt{MountainCar}.

\paragraph{Waiting enables downstream interleaving.}
We now describe a simple form of \textit{policy interleaving} as a downstream application of learning to wait. 
Recall from~\Cref{fig:scratch-and-adapt} that our approach adapts a handwritten base policy that completes the \texttt{Coffee} task (by first turning on the coffee machine, then brewing coffee, then collecting sugar and cream) in 54 timesteps into a waiting policy that completes the task in the same amount of timesteps, but which commits to waiting. %
We now assume access to two additional handwritten policies, one for fetching sugar and another for fetching cream. 
Each of these auxiliary policies, when started from a location on the grid, brings the agent to collect its respective item, and returns to the grid position from which the auxiliary policy was invoked. We also assume access to accurate predictions of how long the auxiliary task will take to execute.

We then interleave as follows:
if the waiting policy selects a duration-$N$ wait, we check the time-to-completion estimates for each not-yet-completed auxiliary task and execute the longest-duration auxiliary task that can be completed in $N$ timesteps.
Our waiting wrapper policy's duration-15 waiting gap is long enough to fetch the sugar; performing this interleaved behavior results allows the agent to complete \texttt{Coffee} in a total of 45 timesteps, matching the amount of time taken to complete the task by vanilla Q-Learning applied to learning how to maximize task performance from scratch (\Cref{fig:cluster}).

\section{Related Work}\label{sec:related}
\paragraph{Cost of Agent Intervention.} 
Prior research~\citep{hansen1996reinforcement,heemels2012introduction,nagahara2016maximum,zhou2024timing,krishna2025vosi} demonstrates how imposing costs on sensing and intervention compel the agent to adopt open-loop strategies in which the agent commits to a sequence of actions without intermediate feedback, typically by relying on internal model of the environment until uncertainty necessitates sensing. 
Closest in spirit to our work is~\citet{zhou2024timing}, in which the agent selects both an action and a number of times to repeat the action, although they do not explore how adjusting the magnitude of their scalar \textit{interaction cost} trades off between task performance and policy query frequency.
While we also encourage the agent to sense and deliberate infrequently,
we focus on a stricter notion of non-intervention formalized by the distinguished \textit{wait} action. 
Committing to taking the wait action, i.e., surrendering control to the system's passive dynamics, is qualitatively different than committing to take a predetermined sequence of arbitrary actions because (1.) waiting may conserve motor resources (e.g. the energy needed to walk through a kitchen) and (2.) periods of waiting can be ``filled in'' with a policy that pursues an auxiliary goal, under appropriate conditions on the dynamics under the waiting action and the auxiliary goal policy.

\paragraph{Temporal Abstractions.} RL has long utilized temporal abstractions to improve efficiency in complex domains~\citep{sutton1999between,pateria2021hrlsurvey,hutsebaut2022hierarchical}. 
Grounded in bounded rationality~\citep{simon1957behavioral}, \citet{harb2018when} argue that agents should identify temporally extended options that effectively minimize the \textit{deliberation cost} associated with option selection. 
This is similar in spirit to \Cref{eqn:objective}.
In particular,~\citet{harb2018when} seek to limit the number of times a high-level policy over options may be queried 
and propose a scalarized objective based on the Lagrangian formulation (analogous to our scalarized RL baseline~\cref{sec:scalar}) of this constrained optimization problem.
However, a drawback compared to our proposed approach based on lexicographic RL is the need for extensive hyperparameter sweeps to identify appropriate weighting of deliberation cost (which corresponds to the $\lambda$ value of our scalar approach) relative to the \taskreturn{}.
\textit{Hierarchical RL with Timed Subgoals}~\citep{gurtler2021hierarchical} introduces fine-grained temporal abstraction by letting a high-level policy call \textit{specify when each call to a low-level policy should terminate}.
This is similar in spirit to dynamically-determined action repeat~\citep{lakshminarayanan2017dynamic,sharma2017learning,biedenkapp2021temporl} where, in addition to choosing a primitive action, the agent chooses a number timesteps to repeat that primitive action. 
While these approaches reduce decision frequency in practice, 
their focus is reducing the 
amount of environmental interaction (sample complexity)
needed to train a performant policy. 
This contrasts with our objective, which explicitly seeks to minimize the number of decisions the agent makes during deployment and privileges extended \textit{waiting} as opposed to extended application of arbitrary primitive actions.

\paragraph{Multi-objective RL.}
Multi-objective MDPs~(MOMDPs) are often employed when a single scalar reward cannot capture the trade-offs between competing desiderata~\citep{roijers2013survey,hayes2022practical}. While optimizing a scalarized objective can make use of standard RL techniques, the objective itself requires precise tuning of weights to navigate the Pareto front. 
In contrast, lexicographic approaches~\citep{gabor1998multi,skalse2022lexicographic} allow for a strict prioritization of objectives, which is well-suited for our setting in which we wish to wait as much as possible without sacrificing expected \taskreturn{}.
This is closely related to the motivation of constrained policy optimization in safe RL~\citep{garcia2015comprehensive,gu2024review}; however, rather than choose an \textit{a priori} constraint on the minimal acceptable task performance the lexicographic approach allows the agent to adaptively discover what the best possible task performance and optimize waiting subject to this implicit constraint.

\section{Discussion}\label{sec:discussion}
\paragraph{Summary.} 
In this paper we formulated learning to wait as maximizing~\Cref{eqn:objective}, which encourages the agent to spend long durations committed to apply the \textit{wait} action without sensing, to the extent possible without sacrificing expected \taskreturn{}.
Lexicographic MORL allows the agent to discover the optimal expected \taskreturn{}, alleviating the need for a human to perform hyperparameter sweeps over a reward weighting coefficient.
We can thus train a policy, or a wrapper around an existing policy, that waits as much as possible without sacrificing task performance. Waiting inherently saves sensory, computational, and motor resources and may be a useful first step towards multi-task policy interleaving.

\paragraph{Future work.} Currently, our approach does not leverage the inherent \textit{structure} of extended duration waiting.
Future work could improve sample efficiency during training by e.g. fitting a single-step dynamics model to multi-timestep waiting experiences, taking inspiration from~\citet{zhou2024timing}.
Secondly, we only expect our techniques to work well in \textit{fully observed} environments.
If, for example, the agent's observations did not differentiate when the soup in the \texttt{Cook} task was just starting to cook vs. when it was almost ready, the agent would not be able to learn appropriate waiting behavior.
An exciting direction of future work would be to learn \textit{memoryful} waiting policies. In many practical settings (e.g. cooking) the task-relevant latent dynamics are highly correlated with the passage of time; which may permit accurate deterministic belief updates.
Finally, we presented only an initial proof-of-concept case study of policy interleaving.
To truly expose opportunities to perform auxiliary tasks would require a stringent definition of what actions cause the aspects of the environment's state that are relevant to the primary task to evolve as they would under the waiting action. Without such restrictions, the progress made towards the auxiliary task could impede the completion of the primary task.

\subsubsection*{Acknowledgments}
\label{sec:ack}
This project was funded by NSF SLES 2331783, NSF CAREER 2239301, ONR N00014-22-1-2677, DARPA TIAMAT HR00112490421, and a gift from Amazon AWS to the ASSET center at Penn.

\bibliography{main}
\bibliographystyle{rlj}

\beginSupplementaryMaterials
\appendix

\section{Additional Environment Details}\label{app:env-details}
\paragraph{Cook.} The \texttt{Cook}, \texttt{CookLonger}, and \texttt{CookTwice}, environments are implemented using~\citet{caroll2019utility}. All three variants use the same kitchen layout shown in~\cref{fig:coffee}. The environment has deterministic 2D grid dynamics: the agent has 6 actions available: go left, go right, go up, go down, stay=$\wait$, or interact. Interact will pick up items (onion, bowl, finished soup) or put down the object the agent is holding. In~\Cref{fig:concept} we write ``turn-left'' to mean go left, and ``take-soup'' to mean interact.
The agent's discrete observation space includes the agent's 2D position, orientation, whether the agent is holding an \{onion, dish, ready soup, uncooked soup\}, the pot phase (``empty'', ``has some but not all ingredients'', ``is cooking'', or ``is done''), how many ingredients are in the pot, and the time remaining for a currently cooking soup to be done. The maximum episode horizon $H$ is 200 for each variant.

\paragraph{Coffee.}
The \texttt{Coffee} environment is a 2D gridworld implemented using Minigrid~\citep{chevalier2023minigrid}. The discrete state is defined by the agent's position, orientation (north, south, east, or west), whether the sugar has been collected or not, whether the cream has been collected or not, the progress of the coffee machine towards heating up (not toggled yet, the number of timesteps remaining to heat, or that it is fully heated), the progress of the coffee machine towards brewing (not toggled yet, the number of timesteps remaining to brew, or that the coffee is brewed and ready to collect), and whether the coffee has been collected or not.
The agent observes the complete state at each timestep.
We use the standard set of 7 discrete actions available to a Minigrid agent but define custom behavior (indicated in parentheses where appropriate): ``turn left'', ``turn right'', ``go forward'', ``pickup'' (no-op), ``drop'' (no-op), ``toggle'', and ``done''=$\wait$ (we redefine ``done'' to make the agent stay in place with no special effect on episode termination). The ``toggle'' action is used for all interactions with the world (starting coffee machine to warm up, starting coffee machine to brew, collecting coffee, collecting cream, collecting sugar).
The maximum episode horizon is $H=200$.
A truly optimal policy completes \texttt{Coffee} in 41 timesteps, but none of our learning based approaches learned the optimal behavior during our allotted training duration (10m timesteps); our best policies need 45 timesteps.

In our interleaving case study, the ``fetch sugar'' and ``fetch cream'' sub-policies are handwritten optimal policies that, when called for a particular agent position and orientation, make the agent go to the sugar (resp. cream) location, ``toggle'' to pick up the sugar (resp. cream), and return the agent to the position and orientation from which the subpolicy was called.
It might seem counterintuitive that our simple scheduler does not choose to execute the ``fetch cream'' policy during the duration-5 wait that the wrapper waiting policy (wrapped around our handwritten policy for the entire \texttt{Coffee} task) executes while waiting for the coffee machine to heat up.
This is because it would take more than 5 timesteps to collect the cream \textit{and return to the position and orientation from the start of the duration-5 wait.}
Overall task completion would be faster if the agent were able to fetch the cream while the coffee machine heats up; future work could explore more sophisticated forms of policy interleaving to enable such behavior.

\paragraph{Pong.}
We adapt Gymnax's~\citep{gymnax2022github} implementation of the well-known Pong environment.
The agent's primitive action space $\mathcal{A}$ comprises four actions: $\texttt{move\_paddle\_up}$ $\texttt{move\_paddle\_down}$, $\wait$ (which leaves the paddle in place), and $\texttt{hit}$ which leaves the paddle in place and returns the ball if contact would be made. We include the nonstandard \texttt{hit} action to make the connection between waiting and not actively affecting the environment thematically clear; our techniques would work equally well with standard pong dynamics.
The goal is to score one point as fast as possible against a scripted opponent; the maximum episode horizon is $H=1000$.
Our \texttt{Pong} environment includes the standard ``sticky actions'' introduced by~\citet{machado2018revisiting} to introduce a small amount of stochasticity into transitions. This makes the environment very slightly non-Markovian with respect to the observations supplied to the agent.

\paragraph{CartPole.}
We adapt the standard \texttt{CartPole-v0}~\citep{barto1983neuronlike} environment from Gymnasium~\citep{towers2024gymnasium} to have an additional $\wait$ action that applies no force to the cart. Thus the action set is \{``left'', ``right'', $\wait$\}.
The maximum episode horizon is $H=200$ and the task reward is $+1$ for each step.
We retain the typical termination logic from the original implementation.

\paragraph{MountainCar.} We use the standard \texttt{MountainCar-v0}~\citep{moore90efficientmemory-based} environment implementation from Gymnasium.
There are three primitive actions: ``accelerate to the left'', ``don't accelerate''=$\wait$, and ``accelerate to the right.''
The task reward is -1 per timestep and the maximum episode horizon is $H=200.$

\section{Training Details}\label{app:training}
During training, we terminate training episodes once the \textit{a priori} episode horizon is reached or (in the case of goal-reaching tasks) when the goal is reached. For \texttt{CartPole} we inherit the usual termination behavior when the pole's angle becomes excessively low.

For each of \texttt{Cook}, \texttt{CookLonger},  \texttt{CookTwice}, and \texttt{Coffee} we train each policy for 10 million environment steps.
For each of our continuous state tasks we train each policy for 1 million timesteps.

When selecting the best model checkpoint of a training run, we use the environment's cumulative reward (for \lexq{} and \ldqn{} we use the lexicographic ordering over the task and waiting rewards, for the other approaches we use the usual ordering over the (scalarized or vanilla) reward). We evaluate an intermediate checkpoint every 10k training environment steps, and evaluate 20 eval episodes (using greedy, not epsilon-greedy policy inference) to obtain an empirical estimate of the expected cumulative reward.

We report the hyperparameters used for all experiments. Lexicographic Q-learning parameters are shown in Table~\ref{tab:lex-qlearning-params}, and lexicographic DQN parameters are shown in Table~\ref{tab:lex-dqn-params}. DQN hyperparameters were adapted from the RL Baselines3 Zoo~\citep{rl-zoo3} and manually tuned for each environment.

\subsection{Lexicographic RL Implementation}\label{app:ldqn}
We adapt the Q-Learning and DQN implementations from SKRL~\citep{serrano2023skrl} into the \lexq{} and \ldqn{} algorithms following~\citet{skalse2022lexicographic}.
We also add support for durative macro-actions as described for the \textit{SMDP Q-Learning} algorithm in~\citet{sutton1999between}.
Durative macro-actions were not part of the standard MOMDP environments treated by~\citet{skalse2022lexicographic} or the base implementation of SKRL.

Our Q-Learning and \lexq{} implementations do not make use of a replay buffer.
Our DQN and \ldqn{} implementations do use a replay buffer: we insert durative-action transitions into the replay buffer in the straightforward way, that is, we include the state from which the action was taken, the (durative) action, the state reached after finishing (or reaching episode termination) the durative action, and the cumulative reward accrued during the low-level transitions within the durative action.

\begin{table}[htbp]
\centering
\caption{\lexq{} hyperparameters.}
\label{tab:lex-qlearning-params}
\begin{tabular}{@{}lcc@{}}
\toprule
\textbf{Parameter} & \textbf{Symbol} & \textbf{Value} \\
\midrule
Discount factor & $\gamma$ & $1.0$ \\
Learning rate & $\alpha$ & $0.1$ \\
Exploration rate & $\varepsilon$ & $0.05$ \\
Lexicographic slack & $\sigma$ & $0.001$ \\
\bottomrule
\end{tabular}
\end{table}

\begin{table}[htbp]
\centering
\caption{Lexicographic DQN hyperparameters by environment.}
\label{tab:lex-dqn-params}
\begin{tabular}{@{}lccc@{}}
\toprule
\textbf{Parameter} & \textbf{CartPole} & \textbf{MountainCar} & \textbf{Pong} \\
\midrule
\multicolumn{4}{@{}l}{\textit{Network Architecture}} \\
\quad Hidden layers & $[256, 256]$ & $[256, 256]$ & $[256, 256]$ \\
\quad Activation & ReLU & ReLU & ReLU \\
\midrule
\multicolumn{4}{@{}l}{\textit{Core Hyperparameters}} \\
\quad Discount factor ($\gamma$) & $1.0$ & $1.0$ & $1.0$ \\
\quad Learning rate ($\alpha$) & $2 \times 10^{-3}$ & $4 \times 10^{-3}$ & $4 \times 10^{-4}$ \\
\quad Batch size & $64$ & $128$ & $32$ \\
\midrule
\multicolumn{4}{@{}l}{\textit{Replay Buffer}} \\
\quad Buffer size & $100{,}000$ & $10{,}000$ & $100{,}000$ \\
\midrule
\multicolumn{4}{@{}l}{\textit{Training Schedule}} \\
\quad Learning starts & $1{,}000$ & $1{,}000$ & $1{,}000$ \\
\quad Gradient steps & $128$ & $8$ & $1$ \\
\quad Update interval & $256$ & $16$ & $4$ \\
\midrule
\multicolumn{4}{@{}l}{\textit{Target Network}} \\
\quad Update interval & $10$ & $600$ & $1{,}000$ \\
\quad Polyak ($\tau$) & $1.0$ & $1.0$ & $1.0$ \\
\midrule
\multicolumn{4}{@{}l}{\textit{Exploration ($\varepsilon$-greedy)}} \\
\quad Initial $\varepsilon$ & $1.0$ & $1.0$ & $1.0$ \\
\quad Final $\varepsilon$ & $0.04$ & $0.07$ & $0.01$ \\
\quad Decay timesteps & $8{,}000$ & $24{,}000$ & $100{,}000$ \\
\bottomrule
\end{tabular}
\end{table}

\section{Learning Curves}\label{app:curves}
We now present learning curves for the training runs reported in~\cref{fig:scratch-and-adapt} and \Cref{fig:cluster}. Note that~\cref{fig:cluster} presents just the best (in terms~\cref{eqn:objective} for \lexq{} and \ldqn{} and in terms of scalarized return for scalarized approaches) policy checkpoint, which may appear before the max number of training environment steps is reached.
In the learning curves, the bold line shows the average of ten training seeds. The shaded region shows 1 standard deviation. 
For some environments, notably \texttt{CookTwice}, policy optimization is unstable, meaning that averaging performance across ten random seeds does not clearly reflect the performance of each individual seed's policy.

For all approaches, we provide learning curves for episode length (negative $J^0$ for most environments; positive $J^0$ for \texttt{CartPole}). This characterizes the cumulative task reward for our lexicographic MORL and the cumulative reward for vanilla RL.
For all approaches except vanilla RL, we also include a curve for number of decisions taken (negative $J^1$).
This characterizes the cumulative waiting reward for \lexq{} and \ldqn{} approaches.
For the scalarized approaches we also include a learning curve for the scalarized return, which is always a linear combination of $J^0$ and $J^1$ weighted by the coefficient $\lambda$ as described in~\cref{sec:scalar}.

\begin{figure}[!h]
    \centering

     \includegraphics[width=\columnwidth]{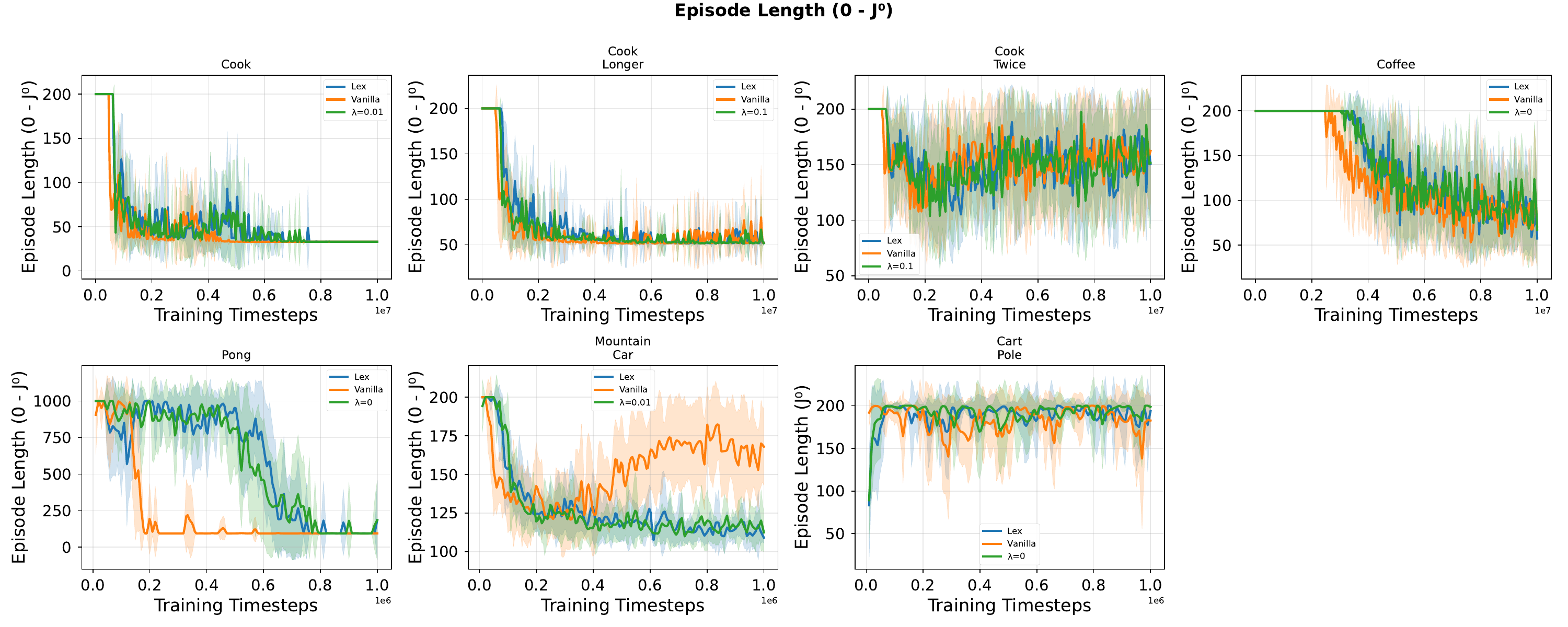}

    \caption{Episode length vs. training timesteps for learning from scratch experiments (see~\cref{fig:cluster}). Episode length equals negative cumulative task return for most environments and equals positive task return for \texttt{CartPole}. Each curve is the mean of 10 random policy training seeds; shaded regions show $\pm$ 1 standard deviation. Each point along the curve is the empirical mean over 20 evaluation trajectories (with greedy policy inference); evaluation occurs once every 10,000 environment steps during training.}
    \label{fig:learning-curves-1}
    
\end{figure}

\begin{figure}[!h]
    \centering
        \includegraphics[width=\columnwidth]{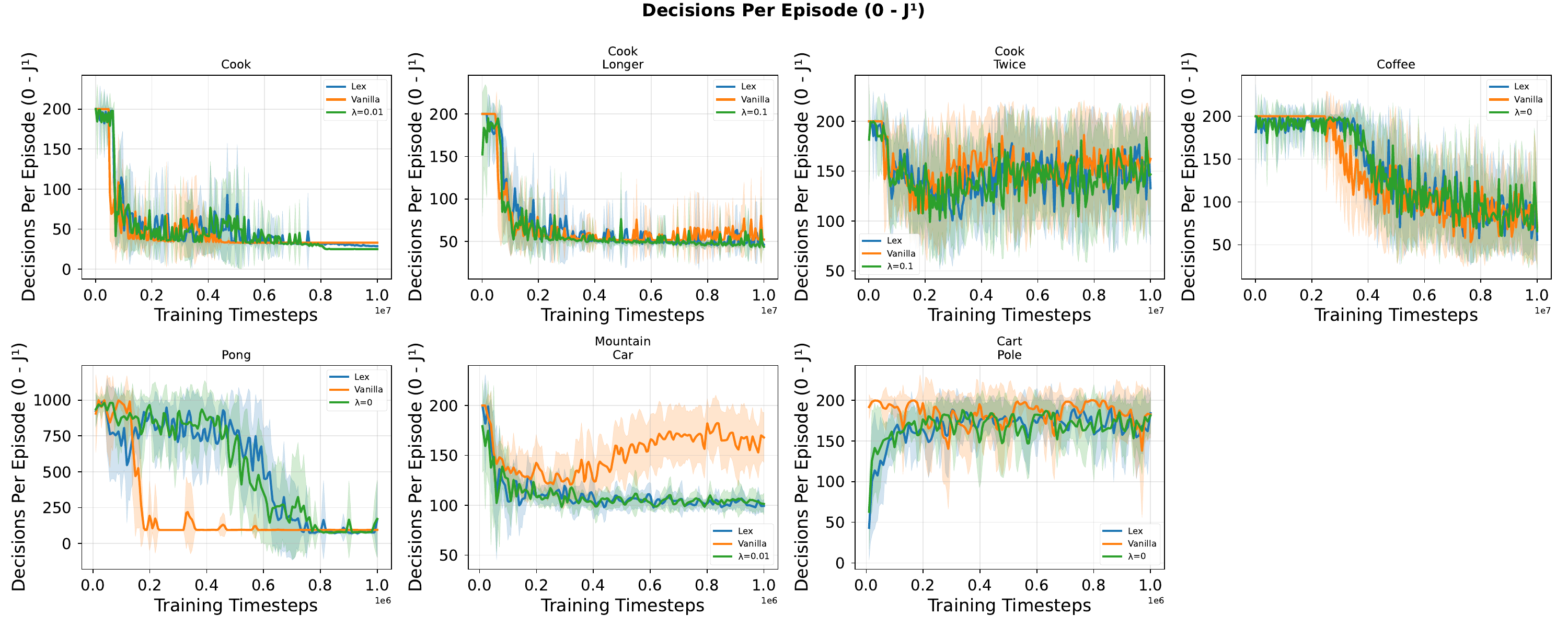}

    \caption{Decisions per episode vs. training timesteps for learning from scratch experiments (see~\cref{fig:cluster}). Decisions per episode is negative cumulative waiting return for all environments. Each curve is the mean of 10 random policy training seeds; shaded regions show $\pm$ 1 standard deviation. Each point along the curve is the empirical mean over 20 evaluation trajectories (with greedy policy inference); evaluation occurs once every 10,000 environment steps during training.}
    \label{fig:learning-curves-2}

\end{figure}
\begin{figure}[!h]
    \centering
    
    \includegraphics[width=\columnwidth]{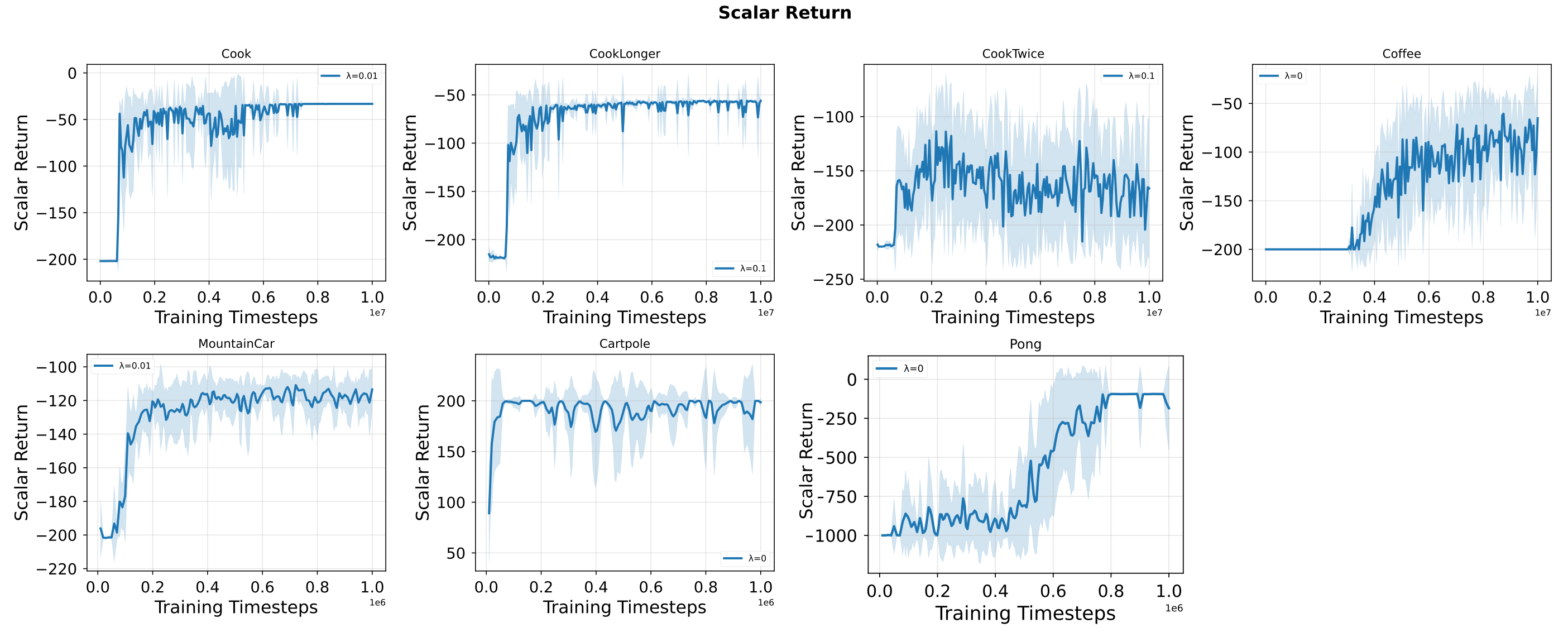}

    \caption{Per-episode cumulative scalar reward vs. training timesteps for learning from scratch experiments (see~\cref{fig:cluster}). Each curve is the mean of 10 random policy training seeds; shaded regions represents $\pm$ 1 standard deviation. Each point along the curve is the empirical mean over 20 evaluation trajectories (with greedy policy inference); evaluation occurs once every 10,000 environment steps during training. We only report curves for the best scalarization coefficient $\lambda$, as described in~\cref{sec:experiments}.}

    \label{fig:learning-curves-scalar}

\end{figure}

\begin{figure}[!h]
    \centering
       \includegraphics[width=\columnwidth]{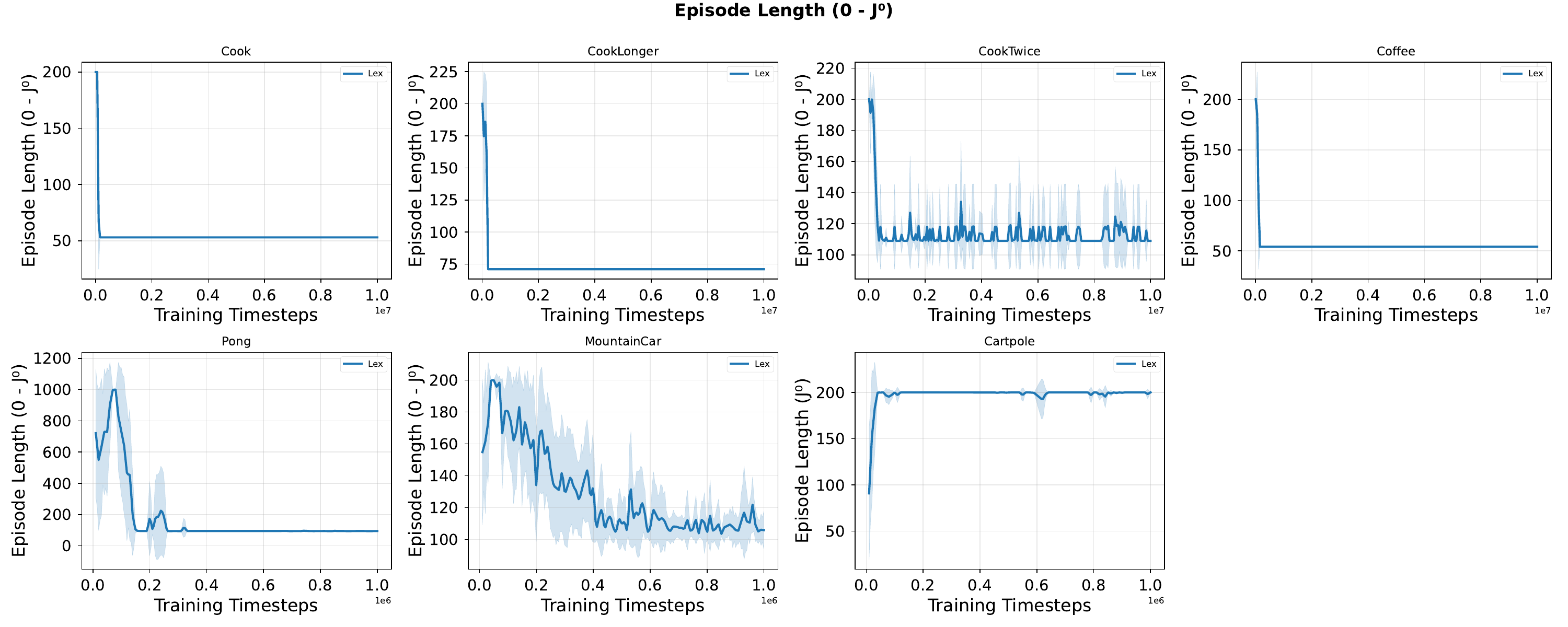}
    \caption{Episode length vs. training timesteps for learning from wrapper waiting policy experiments (see~\cref{fig:scratch-and-adapt}). Episode length equals negative task return for most environments and equals positive task return for \texttt{CartPole}. Each curve is the mean of 10 random policy training seeds; shaded regions represents $\pm$ 1 standard deviation. Each point along the curve is the empirical mean over 20 evaluation trajectories (with greedy policy inference); evaluation occurs once every 10,000 environment steps during training.}
    \label{fig:learning-curves-adapt-1}
    
\end{figure}
\begin{figure}[h]
    \centering

     \includegraphics[width=\columnwidth]{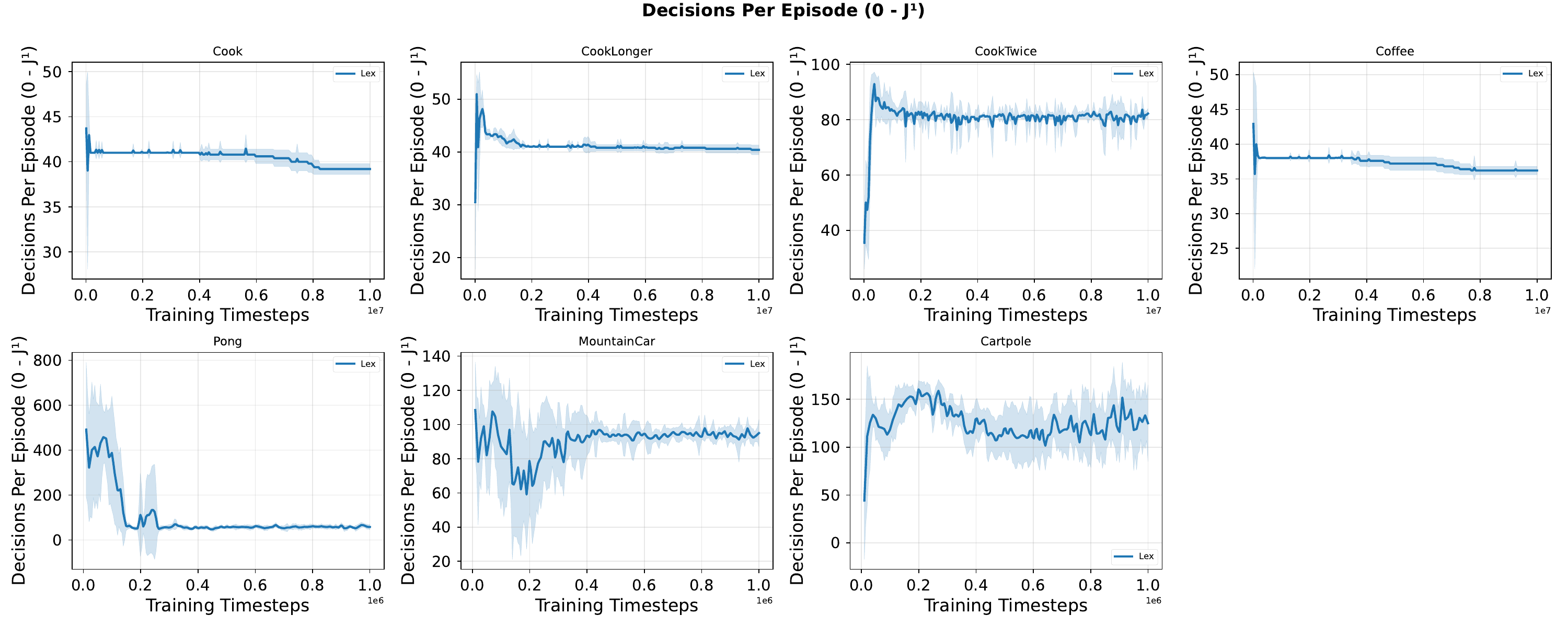}

    \caption{Decisions per episode vs. training timesteps for learning from wrapper waiting policy experiments (see~\cref{fig:scratch-and-adapt}). Decisions per episode is negative waiting return for all environments. Each curve is the mean of 10 random policy training seeds; shaded regions represents $\pm$ 1 standard deviation. Each point along the curve is the empirical mean over 20 evaluation trajectories (with greedy policy inference); evaluation occurs once every 10,000 environment steps during training.}
    \label{fig:learning-curves-adapt-2}

\end{figure}

\FloatBarrier

\subsection{Extended Tabular results}\label{app:extended-results}
In the following results tables, we report the mean and standard deviation for all experiments performed.
This includes scalarized training values of $\lambda$ that do not appear in~\cref{fig:cluster}.
For some environments, we extended our sweep to additional values of $\lambda$. 
Unless otherwise specified, we report mean and standard deviation over 10 training runs. Our results for learning a waiting wrapper policy using scalarized rewards are reported over 5 training runs.

\begin{table}[htbp]
  \centering
  \begin{tabular}{lll}
     Method & Episode Length ($0-J^0$) & Policy Queries per Episode ($0-J^1$) \\
    \hline

      Scratch \lexq{} & 33.0 $\pm$ 0.0 & 25.0 $\pm$ 0.0 \\
      Scratch scalar 0 & 33.0 $\pm$ 0.0 & 32.2 $\pm$ 1.687 \\
      Scratch scalar 0.01 & 33.0 $\pm$ 0.0 & 25.0 $\pm$ 0.0 \\
      Scratch scalar 0.1 & 33.0 $\pm$ 0.0 & 25.0 $\pm$ 0.0 \\
      Scratch scalar 0.5 & 33.0 $\pm$ 0.0 & 25.0 $\pm$ 0.0 \\
      Scratch scalar 1 & 33.0 $\pm$ 0.0 & 25.0 $\pm$ 0.0 \\
      Scratch scalar 5 & 37.0 $\pm$ 0.0 & 23.0 $\pm$ 0.0 \\
      Scratch scalar 10 & 37.0 $\pm$ 0.0 & 23.0 $\pm$ 0.0 \\
      Scratch scalar 15 & 37.0 $\pm$ 0.0 & 23.0 $\pm$ 0.0 \\
      Scratch scalar 20 & 102.2 $\pm$ 84.173 & 19.4 $\pm$ 4.648 \\
    Scratch vanilla Q-Learning & 33.0 $\pm$ 0.0 & 33.0 $\pm$ 0.0 \\
      \midrule
      Wrap \lexq{} & 53.0 $\pm$ 0.0 & 39.0 $\pm$ 0.0 \\
      Wrap scalar 0 & 53.0 $\pm$ 0.0 & 53.0 $\pm$ 0.0 \\
      Wrap scalar 0.01 & 53.0 $\pm$ 0.0 & 39.0 $\pm$ 0.0 \\
      Wrap scalar 0.1 & 53.0 $\pm$ 0.0 & 39.0 $\pm$ 0.0 \\
      Wrap scalar 0.5 & 53.0 $\pm$ 0.0 & 39.0 $\pm$ 0.0 \\
      Wrap scalar 1 & 53.0 $\pm$ 0.0 & 39.0 $\pm$ 0.0 \\
      Wrap scalar 5 & 56.0 $\pm$ 0.0 & 38.0 $\pm$ 0.0 \\
      Wrap scalar 10 & 200.0 $\pm$ 0.0 & 14.2 $\pm$ 0.447 \\
      Wrap scalar 15 & 56.0 $\pm$ 0.0 & 38.0 $\pm$ 0.0 \\
      Wrap scalar 20 & 200.0 $\pm$ 0.0 & 14.2 $\pm$ 0.447 \\
    \hline
  \end{tabular}
  \caption{\texttt{Cook} results. ``Wrap'' refers to wrapping a base policy that is handwritten and which stands idly by as soup cooks.} 
  \label{tab:table-cook-10m}
\end{table}

\begin{table}[htbp]
  \centering

 \begin{tabular}{lll}
     Method & Episode Length ($0-J^0$) & Policy Queries per Episode ($0-J^1$) \\
    \hline
     Scratch \lexq{} & 51.0 $\pm$ 0.0 & 33.8 $\pm$ 1.687 \\
     Scratch scalar 0 & 51.0 $\pm$ 0.0 & 48.2 $\pm$ 1.932 \\
     Scratch scalar 0.01 & 51.0 $\pm$ 0.0 & 34.4 $\pm$ 1.897 \\
     Scratch scalar 0.1 & 51.0 $\pm$ 0.0 & 33.4 $\pm$ 1.265 \\
     Scratch scalar 0.5 & 51.4 $\pm$ 0.966 & 31.2 $\pm$ 2.53 \\
     Scratch scalar 1 & 51.4 $\pm$ 0.966 & 29.8 $\pm$ 1.751 \\
     Scratch scalar 5 & 52.0 $\pm$ 0.0 & 24.0 $\pm$ 0.0 \\
     Scratch scalar 10 & 52.0 $\pm$ 0.0 & 24.0 $\pm$ 0.0 \\
     Scratch scalar 15 & 111.2 $\pm$ 76.427 & 20.0 $\pm$ 5.164 \\
Scratch vanilla Q-Learning & 51.0 $\pm$ 0.0 & 51.0 $\pm$ 0.0 \\
    
    \midrule
Wrap \lexq{} & 71.0 $\pm$ 0.0 & 40.2 $\pm$ 1.033 \\

      Wrap scalar 0 & 71.0 $\pm$ 0.0 & 71.0 $\pm$ 0.0 \\
      Wrap scalar 0.01 & 71.0 $\pm$ 0.0 & 39.0 $\pm$ 0.0 \\
      Wrap scalar 0.1 & 71.0 $\pm$ 0.0 & 39.0 $\pm$ 0.0 \\
      Wrap scalar 0.5 & 71.0 $\pm$ 0.0 & 39.0 $\pm$ 0.0 \\
      Wrap scalar 1 & 71.0 $\pm$ 0.0 & 39.0 $\pm$ 0.0 \\
      Wrap scalar 5 & 71.0 $\pm$ 0.0 & 39.0 $\pm$ 0.0 \\
      Wrap scalar 10 & 200.0 $\pm$ 0.0 & 14.2 $\pm$ 0.447 \\
      Wrap scalar 15 & 200.0 $\pm$ 0.0 & 14.0 $\pm$ 0.0 \\
    \hline
  \end{tabular}
  \caption{\texttt{CookLonger} results. ``Wrap'' refers to wrapping a base policy that is handwritten and which stands idly by as soup cooks.} 
  \label{tab:table-cooklonger-10m}
\end{table}

\begin{table}[htbp]
  \centering
   \begin{tabular}{lll}
     Method & Episode Length ($0-J^0$) & Policy Queries per Episode ($0-J^1$) \\
    \hline

     Scratch \lexq{} & 69.0 $\pm$ 0.0 & 64.2 $\pm$ 4.131 \\
     Scratch scalar 0 & 69.2 $\pm$ 0.632 & 68.4 $\pm$ 2.675 \\
     Scratch scalar 0.01 & 69.0 $\pm$ 0.0 & 65.0 $\pm$ 4.216 \\
     Scratch scalar 0.1 & 69.0 $\pm$ 0.0 & 64.2 $\pm$ 4.131 \\
     Scratch scalar 0.5 & 69.8 $\pm$ 1.033 & 61.0 $\pm$ 2.309 \\
     Scratch scalar 1 & 71.8 $\pm$ 3.676 & 57.4 $\pm$ 4.502 \\
     Scratch scalar 5 & 200.0 $\pm$ 0.0 & 15.1 $\pm$ 1.287 \\
     Scratch scalar 10 & 200.0 $\pm$ 0.0 & 15.7 $\pm$ 2.946 \\
     Scratch scalar 15 & 200.0 $\pm$ 0.0 & 14.5 $\pm$ 0.527 \\
     Scratch scalar 20 & 200.0 $\pm$ 0.0 & 14.5 $\pm$ 0.527 \\
Scratch vanilla Q-Learning & 69.0 $\pm$ 0.0 & 69.0 $\pm$ 0.0 \\
    \midrule
     Wrap \lexq{} & 109.0 $\pm$ 0.0 & 81.0 $\pm$ 0.0 \\

      Wrap scalar 0 & 109.0 $\pm$ 0.0 & 81.0 $\pm$ 0.0 \\
      Wrap scalar 0.01 & 109.0 $\pm$ 0.0 & 81.0 $\pm$ 0.0 \\
      Wrap scalar 0.1 & 109.0 $\pm$ 0.0 & 81.0 $\pm$ 0.0 \\
      Wrap scalar 0.5 & 109.0 $\pm$ 0.0 & 81.0 $\pm$ 0.0 \\
      Wrap scalar 1 & 200.0 $\pm$ 0.0 & 14.0 $\pm$ 0.0 \\
      Wrap scalar 5 & 200.0 $\pm$ 0.0 & 14.0 $\pm$ 0.0 \\
      Wrap scalar 10 & 200.0 $\pm$ 0.0 & 14.0 $\pm$ 0.0 \\
      Wrap scalar 15 & 200.0 $\pm$ 0.0 & 14.0 $\pm$ 0.0 \\
      Wrap scalar 20 & 200.0 $\pm$ 0.0 & 14.0 $\pm$ 0.0 \\
    \hline
  \end{tabular}
  \caption{\texttt{CookTwice} results. ``Wrap'' refers to wrapping a base policy that is handwritten and which stands idly by as soup cooks.} 
  \label{tab:table-cooktwice-10m}
\end{table}
\begin{table}[htbp]
  \centering
 \begin{tabular}{lll}
     Method & Episode Length ($0-J^0$) &Policy Queries per Episode ($0-J^1$) \\
    \hline
     Scratch \lexq{} & 45.5 $\pm$ 0.707 & 45.1 $\pm$ 1.595 \\
      Scratch scalar 0 & 45.1 $\pm$ 0.568 & 44.7 $\pm$ 1.418 \\
      Scratch scalar 0.01 & 45.3 $\pm$ 0.823 & 44.5 $\pm$ 1.179 \\
      Scratch scalar 0.1 & 45.5 $\pm$ 0.972 & 45.1 $\pm$ 1.101 \\
      Scratch scalar 0.5 & 45.4 $\pm$ 0.843 & 41.0 $\pm$ 2.494 \\
      Scratch scalar 1 & 45.75 $\pm$ 1.165 & 40.5 $\pm$ 3.295 \\
      Scratch scalar 2 & 47.6 $\pm$ 2.797 & 38.0 $\pm$ 2.211 \\
      Scratch scalar 5 & 49.0 $\pm$ 3.162 & 36.429 $\pm$ 1.272 \\
      Scratch scalar 10 & 156.1 $\pm$ 70.699 & 21.1 $\pm$ 9.231 \\
      Scratch scalar 15 & 200.0 $\pm$ 0.0 & 14.4 $\pm$ 0.699 \\
      Scratch scalar 20 & 200.0 $\pm$ 0.0 & 14.2 $\pm$ 0.422 \\
      Scratch scalar 30 & 200.0 $\pm$ 0.0 & 14.8 $\pm$ 1.229 \\
     Scratch vanilla Q-Learning & 44.8 $\pm$ 0.632 & 44.8 $\pm$ 0.632 \\

     \midrule
      Wrap \lexq{} & 54.0 $\pm$ 0.0 & 36.0 $\pm$ 0.0 \\
      
      Wrap scalar 0 & 54.0 $\pm$ 0.0 & 54.0 $\pm$ 0.0 \\
      Wrap scalar 0.01 & 54.0 $\pm$ 0.0 & 36.0 $\pm$ 0.0 \\
      Wrap scalar 0.1 & 54.0 $\pm$ 0.0 & 36.0 $\pm$ 0.0 \\
      Wrap scalar 0.5 & 54.0 $\pm$ 0.0 & 36.0 $\pm$ 0.0 \\
      Wrap scalar 1 & 54.0 $\pm$ 0.0 & 36.0 $\pm$ 0.0 \\
      Wrap scalar 2 & 54.0 $\pm$ 0.0 & 36.0 $\pm$ 0.0 \\
      Wrap scalar 5 & 57.0 $\pm$ 0.0 & 35.0 $\pm$ 0.0 \\
      Wrap scalar 10 & 200.0 $\pm$ 0.0 & 14.0 $\pm$ 0.0 \\
      Wrap scalar 15 & 200.0 $\pm$ 0.0 & 14.0 $\pm$ 0.0 \\
      Wrap scalar 20 & 200.0 $\pm$ 0.0 & 14.4 $\pm$ 0.894 \\
      Wrap scalar 30 & 200.0 $\pm$ 0.0 & 14.2 $\pm$ 0.447 \\
    \hline
  \end{tabular}
  \caption{\texttt{Coffee} Results. ``Wrap'' refers to wrapping a base policy that is handwritten and which stands idly by as the coffe machine heats and as coffee brews.}
  \label{tab:table-coffee}
\end{table}
\begin{table}[htbp]
  \centering
 \begin{tabular}{lll}
     Method & Episode Length ($0-J^0$) & Policy Queries per Episode ($0-J^1$) \\
    \hline

 Scratch  LDQN & 94.414 $\pm$ 1.091 & 46.898 $\pm$ 9.105 \\
     Scratch scalar 0 & 94.0 $\pm$ 0.0 & 83.929 $\pm$ 14.225 \\
     Scratch scalar 0.01 & 94.09 $\pm$ 0.242 & 45.146 $\pm$ 10.493 \\
     Scratch scalar 0.05 & 94.227 $\pm$ 0.254 & 45.606 $\pm$ 8.269 \\
      Scratch scalar 0.1 & 94.664 $\pm$ 1.523 & 44.793 $\pm$ 7.438 \\
      Scratch scalar 0.2 & 94.1 $\pm$ 0.217 & 46.874 $\pm$ 3.954 \\
      Scratch scalar 0.5 & 94.083 $\pm$ 0.224 & 33.915 $\pm$ 4.048 \\
      Scratch scalar 1 & 94.208 $\pm$ 0.368 & 26.525 $\pm$ 2.934 \\
      Scratch scalar 2 & 94.449 $\pm$ 0.508 & 22.517 $\pm$ 2.687 \\
      Scratch scalar 5 & 95.203 $\pm$ 1.67 & 22.515 $\pm$ 4.199 \\
      Scratch scalar 10 & 94.694 $\pm$ 1.309 & 25.176 $\pm$ 4.199 \\
      Scratch scalar 15 & 118.437 $\pm$ 30.631 & 25.326 $\pm$ 4.026 \\
      Scratch scalar 20 & 94.025 $\pm$ 0.06 & 24.42 $\pm$ 2.995 \\
      Scratch scalar 30 & 102.392 $\pm$ 22.485 & 27.073 $\pm$ 3.162 \\
      Scratch scalar 50 & 564.16 $\pm$ 459.882 & 15.403 $\pm$ 8.02 \\
      Scratch scalar 100 & 1000.0 $\pm$ 0.0 & 8.2 $\pm$ 0.632 \\
      Scratch vanilla DQN & 94.0 $\pm$ 0.0 & 94.0 $\pm$ 0.0 \\    
      \midrule
           Wrap LDQN & 94.439 $\pm$ 0.415 & 30.499 $\pm$ 3.497 \\

           Wrap scalar 0 & 94.399 $\pm$ 0.231 & 91.586 $\pm$ 6.197 \\
      Wrap scalar 0.01 & 94.401 $\pm$ 0.286 & 32.373 $\pm$ 2.655 \\
      Wrap scalar 0.05 & 94.409 $\pm$ 0.344 & 29.72 $\pm$ 1.54 \\
      Wrap scalar 0.1 & 94.332 $\pm$ 0.477 & 29.733 $\pm$ 2.613 \\
      Wrap scalar 0.2 & 94.363 $\pm$ 0.329 & 27.433 $\pm$ 2.794 \\
      Wrap scalar 0.5 & 94.455 $\pm$ 0.286 & 23.748 $\pm$ 1.966 \\
      Wrap scalar 1 & 94.744 $\pm$ 0.271 & 21.842 $\pm$ 0.864 \\
      Wrap scalar 10 & 96.362 $\pm$ 2.502 & 20.559 $\pm$ 0.544 \\
      Wrap scalar 2 & 95.199 $\pm$ 1.779 & 20.469 $\pm$ 0.727 \\
      Wrap scalar 5 & 96.778 $\pm$ 1.526 & 20.264 $\pm$ 0.748 \\
      Wrap scalar 15 & 281.496 $\pm$ 401.745 & 30.734 $\pm$ 15.112 \\
      Wrap scalar 20 & 287.182 $\pm$ 399.216 & 21.642 $\pm$ 8.706 \\
      Wrap scalar 30 & 1000.0 $\pm$ 0.0 & 8.0 $\pm$ 0.0 \\
      Wrap scalar 50 & 1000.0 $\pm$ 0.0 & 8.4 $\pm$ 0.548 \\
      Wrap scalar 100 & 1000.0 $\pm$ 0.0 & 8.0 $\pm$ 0.0 \\
    \hline
  \end{tabular}
  \caption{\texttt{Pong} results. ``Wrap'' refers to wrapping a base policy that is trained using vanilla DQN.}
  \label{tab:table-pong}
\end{table}
\begin{table}[htbp]
  \centering
  \begin{tabular}{lll}
    \hline
    Method & Episode Length $(J^0)$ & Policy Queries per Episode ($0-J^1$)\\
    \hline

      Scratch LDQN & 199.986 $\pm$ 0.042 & 99.76 $\pm$ 23.421 \\
      Scratch scalar 0 & 200.0 $\pm$ 0.0 & 180.72 $\pm$ 25.435 \\
     Scratch scalar 0.01 & 199.952 $\pm$ 0.106 & 83.298 $\pm$ 28.076 \\
      Scratch scalar 0.1 & 199.846 $\pm$ 0.487 & 81.455 $\pm$ 26.969 \\
      Scratch scalar 0.5 & 199.647 $\pm$ 1.116 & 66.388 $\pm$ 19.676 \\
      Scratch scalar 1 & 199.843 $\pm$ 0.345 & 64.69 $\pm$ 24.025 \\
      Scratch scalar 2 & 162.941 $\pm$ 71.322 & 49.522 $\pm$ 29.661 \\
      Scratch scalar 5 & 128.33 $\pm$ 66.645 & 23.113 $\pm$ 18.95 \\
      Scratch scalar 10 & 145.52 $\pm$ 55.1 & 21.197 $\pm$ 15.645 \\
      Scratch scalar 15 & 41.68 $\pm$ 1.042 & 2.535 $\pm$ 0.067 \\
      Scratch scalar 20 & 41.135 $\pm$ 0.37 & 2.498 $\pm$ 0.019 \\
      Scratch vanilla DQN  & 199.727 $\pm$ 0.542 & 199.727 $\pm$ 0.542 \\
      
       \midrule
Wrap LDQN  & 199.973 $\pm$ 0.084 & 55.359 $\pm$ 11.093 \\
       Wrap scalar 0 & 200.0 $\pm$ 0.0 & 134.08 $\pm$ 36.439 \\
       Wrap scalar 0.01 & 199.824 $\pm$ 0.368 & 52.035 $\pm$ 7.038 \\
       Wrap scalar 0.1 & 200.0 $\pm$ 0.0 & 56.029 $\pm$ 7.818 \\
       Wrap scalar 0.5 & 199.959 $\pm$ 0.093 & 52.433 $\pm$ 11.197 \\
       Wrap scalar 1 & 200.0 $\pm$ 0.0 & 55.852 $\pm$ 7.065 \\
       Wrap scalar 2 & 197.424 $\pm$ 5.76 & 52.249 $\pm$ 5.758 \\
       Wrap scalar 5 & 199.087 $\pm$ 1.577 & 26.8 $\pm$ 4.069 \\
       Wrap scalar 10 & 174.452 $\pm$ 9.828 & 15.105 $\pm$ 1.088 \\
       Wrap scalar 15 & 41.347 $\pm$ 0.308 & 2.512 $\pm$ 0.018 \\
       Wrap scalar 20 & 41.409 $\pm$ 0.27 & 2.518 $\pm$ 0.03 \\
    \hline
  \end{tabular}
  \caption{\texttt{Cartpole} Results. ``Wrap'' refers to wrapping a base policy that was trained using vanilla DQN.}
  \label{tab:table-cartpole}
\end{table}

\begin{table}[htbp]
  \centering
 \begin{tabular}{lll}
     Method & Episode Length ($0-J^0$) & Policy Queries per Episode ($0-J^1$) \\
    \hline
Scratch LDQN  & 103.577 $\pm$ 1.631 & 97.532 $\pm$ 4.436 \\
     Scalar 0 & 103.626 $\pm$ 3.505 & 98.212 $\pm$ 3.246 \\
      Scalar 0.01 & 102.601 $\pm$ 2.841 & 97.978 $\pm$ 2.771 \\
      Scalar 0.1 & 105.242 $\pm$ 3.447 & 96.311 $\pm$ 3.548 \\
      Scalar 0.5 & 107.898 $\pm$ 4.655 & 88.284 $\pm$ 4.549 \\
      Scalar 1 & 125.611 $\pm$ 27.028 & 78.364 $\pm$ 24.184 \\
      Scalar 2 & 200.0 $\pm$ 0.0 & 10.613 $\pm$ 1.627 \\
      Scalar 5 & 200.0 $\pm$ 0.0 & 10.101 $\pm$ 0.318 \\
      Scalar 10 & 200.0 $\pm$ 0.0 & 10.232 $\pm$ 0.518 \\
      Scalar 15 & 200.0 $\pm$ 0.0 & 10.054 $\pm$ 0.131 \\
      Scalar 20 & 200.0 $\pm$ 0.0 & 10.117 $\pm$ 0.343 \\
     Scratch vanilla DQN & 102.791 $\pm$ 5.985 & 102.791 $\pm$ 5.985 \\
   
      \midrule
    Wrap LDQN & 101.512 $\pm$ 2.908 & 95.693 $\pm$ 3.791 \\

      Wrap scalar 0 & 100.284 $\pm$ 0.314 & 95.194 $\pm$ 3.462 \\
      Wrap scalar 0.01 & 100.559 $\pm$ 0.782 & 94.772 $\pm$ 3.535 \\
      Wrap scalar 0.1 & 101.703 $\pm$ 1.28 & 93.395 $\pm$ 3.979 \\
      Wrap scalar 0.5 & 104.484 $\pm$ 4.043 & 90.104 $\pm$ 7.571 \\
      Wrap scalar 1 & 114.675 $\pm$ 21.038 & 85.823 $\pm$ 19.539 \\
      Wrap scalar 2 & 200.0 $\pm$ 0.0 & 10.395 $\pm$ 0.86 \\
      Wrap scalar 5 & 200.0 $\pm$ 0.0 & 10.033 $\pm$ 0.052 \\
      Wrap scalar 10 & 200.0 $\pm$ 0.0 & 10.011 $\pm$ 0.025 \\
      Wrap scalar 15 & 200.0 $\pm$ 0.0 & 10.044 $\pm$ 0.062 \\
      Wrap scalar 20 & 200.0 $\pm$ 0.0 & 10.013 $\pm$ 0.029 \\
      Wrap scalar 50 & 200.0 $\pm$ 0.0 & 10.0 $\pm$ 0.0 \\
    \hline
    
  \end{tabular}
  \caption{Results for \texttt{MountainCar} environment. ``Wrap'' refers to wrapping a base policy that is trained using vanilla DQN.}
  \label{tab:table-mountaincar}
\end{table}

\end{document}